\pdfoutput=1
\documentclass[11pt]{article}

\usepackage[final]{acl}

\usepackage{times}
\usepackage{latexsym}

\usepackage[T1]{fontenc}
\usepackage[utf8]{inputenc}

\usepackage{microtype}

\usepackage{inconsolata}

\usepackage{graphicx}
\usepackage{amsmath} 
\usepackage{amsfonts, amsmath, amssymb}
\usepackage{subfigure}
\usepackage{cleveref}

\usepackage{xcolor}
\usepackage{amsfonts}       % blackboard math symbols
\usepackage[]{mdframed}
\usepackage{fancyvrb}
\usepackage{fvextra}
\usepackage{booktabs}
\usepackage{diagbox}
\usepackage{multirow}

\definecolor{correct}{RGB}{150, 200, 50}
\definecolor{gold}{RGB}{230, 171, 2}
\definecolor{distractor}{RGB}{116, 112, 179}

\newcommand{\Nv}{({\em v})~}

\title{Do RAG Systems Really Suffer From Positional Bias?}

\author{
\textbf{Florin Cuconasu\textsuperscript{1,2}}\thanks{Work conducted while FC being a research intern at TII.}\thanks{\texttt{cuconasu@diag.uniroma1.it}}, 
 \textbf{Simone Filice\textsuperscript{2}}\\
 \textbf{Guy Horowitz\textsuperscript{2},}
 \textbf{Yoelle Maarek\textsuperscript{2},}
 \textbf{Fabrizio Silvestri\textsuperscript{1}}
\\
\\
 \textsuperscript{1}Sapienza University of Rome,
 \textsuperscript{2}Technology Innovation Institute
}

\begin{document}
\maketitle
\begin{abstract}
Retrieval Augmented Generation enhances LLM accuracy by adding passages retrieved from an external corpus to the LLM prompt.
This paper investigates how positional bias--the tendency of LLMs to weight information differently based on its position in the prompt--affects not only the LLM's capability to capitalize on relevant passages, but also its susceptibility to distracting passages.
Through extensive experiments on three benchmarks, we show how state-of-the-art retrieval pipelines, while attempting to retrieve relevant passages, systematically bring highly distracting ones to the top ranks, with over 60\% of queries containing at least one highly distracting passage among the top-10 retrieved passages. % \flo{I'd add a number from the table with HD}\simo{Not here, it would remain too vague as you would need to first define the distracting effect and what you mean with hard distractors}\flo{Yes, you are right. But, It would be nice if we have a number in the abstract to quantify the effect. Maybe saying just that X\% of queries present highly distracting passages; which can be understood even without defining the distracting effect. "with over 60\% of queries containing at least one highly distracting passage among the top-10 retrieved passages."}. 
As a result, the impact of the LLM positional bias, which in controlled settings is often reported as very prominent by related works, is actually marginal in real scenarios since both relevant and distracting passages are, in turn, penalized. Indeed, our findings reveal that sophisticated strategies that attempt to rearrange the passages based on LLM positional preferences do not perform better than random shuffling.%, challenging assumptions about the practical impact of positional bias in real RAG systems.%Therefore, strategies that attempt to rearrange the passages in the prompt based on LLM-preferred positions are not as effective as expected \flo{I generally like it. I think we could strengthen the conclusion. What do you think reporting that randomly shuffling the passages is as effective as the best positioning? "Our findings reveal that sophisticated strategies that attempt to rearrange the passages based on LLM positional preferences perform no better than random shuffling, challenging assumptions about the practical impact of positional bias in real RAG systems."}.

\end{abstract}

\section{Introduction}

Retrieval Augmented Generation (RAG) improves the factual accuracy of LLMs on knowledge-intensive tasks by including in the prompt passages retrieved from an external corpus \citep{chen2017reading,petroni2021kilt,fan2024survey}.
Because any real retriever is imperfect, RAG systems feed the LLM \emph{several} top-ranked passages, not just the single best one.  
That practice raises recall but also inserts \emph{distracting} passages: text that looks relevant yet lacks the appropriate answer. Recent work shows these distractors can sharply degrade the LLM answer accuracy \citep{cuconasu2024power,jin2025longcontext,yoran2024making}.

A second, orthogonal weakness of LLMs is \emph{positional bias}: moving the same evidence to a different location in the context can change the answer and largely impact its accuracy.  
\citet{liu-etal-2024-lost} term this the \emph{lost-in-the-middle} effect, to refer to the tendency of LLMs to focus on text appearing in the beginning or end of their prompt. %; while follow-up studies confirm that different models exhibit distinct position–accuracy curves \cite{hutter-2025-lost-but-not-only}.  
Prior analyses \cite{liu-etal-2024-lost,hutter-2025-lost-but-not-only,he-etal-2024-never}, however, study the problem in a controlled setting, typically rotating the position of a sole relevant passage in a prompt otherwise containing only irrelevant passages. This artificial configuration not only amplifies the impact of the positional bias but also ignores how the positional bias influences the vulnerability of the LLMs to distracting passages, which instead is central in our work.

Using the ``distracting effect'' metric of \citet{amiraz2025distracting}, we show that answer accuracy depends on the positions of \emph{both} relevant and distracting passages. Then, we empirically show that current state-of-the-art retrieval pipelines, while attempting to retrieve relevant passages, also bring highly distracting passages to the top ranks, and the more advanced the retrieval pipeline is, the more distracting the passages are. 
%ym - not my style to give actual figures in Introduction it's a spoiler, we will say it in conclusion. Also BGE large is  not referred yet. So commenting out
%For example, among the top-10 passages retrieved by BGE large, there is at least a hard distracting one (i.e., having distracting effect higher than 0.8) for more than 60\% of the queries across three QA benchmarks. 
This simultaneous presence of relevant and highly distracting passages near the top of the retrieval ranking drastically reduces the impact of the positional bias, since it penalizes, in turn, both passage types. 

Following these findings, we empirically demonstrate that strategies to rearrange the passages in the prompt based on the LLM-preferred positions are not more effective than a random passage ordering.% the passages. %and don't surpass a random passage ordering. 

\begin{figure*}[ht]
    \centering
    \subfigure[]{\includegraphics[width=0.24\textwidth]{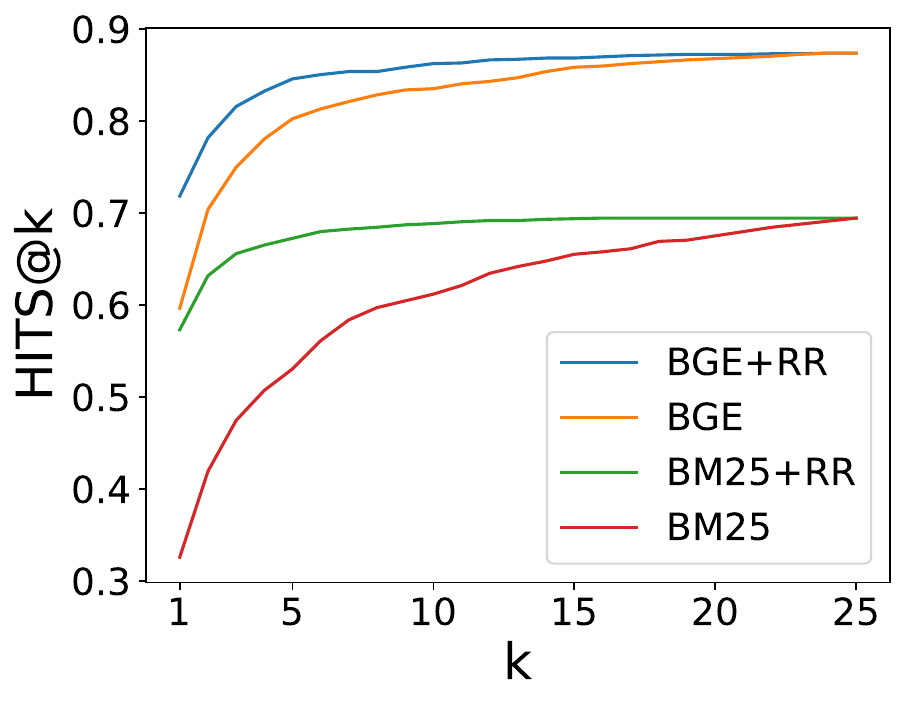}} 
    \subfigure[]{\includegraphics[width=0.24\textwidth]{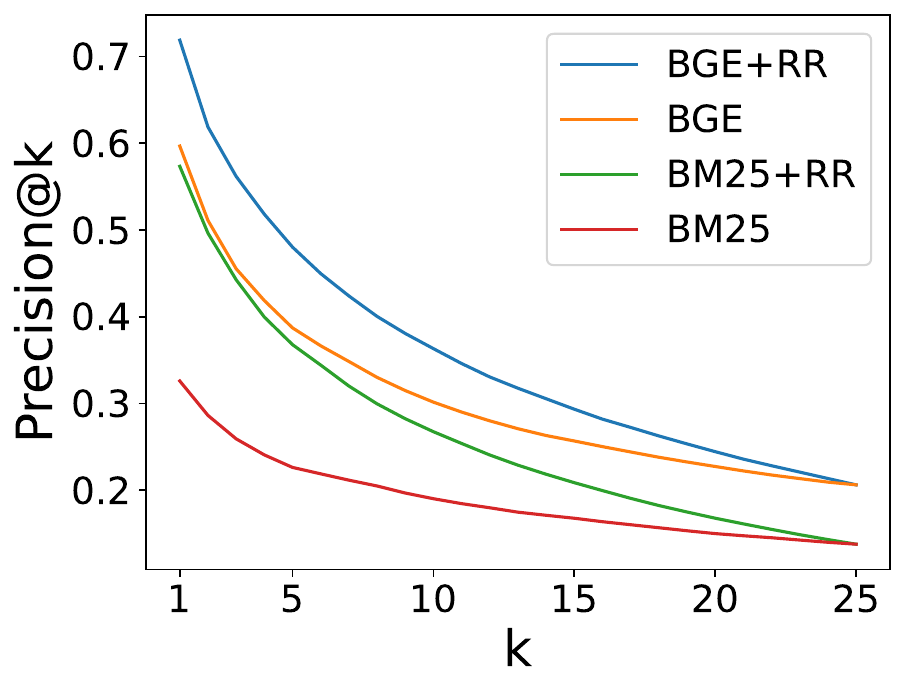}} 
    \subfigure[]{\includegraphics[width=0.24\textwidth]{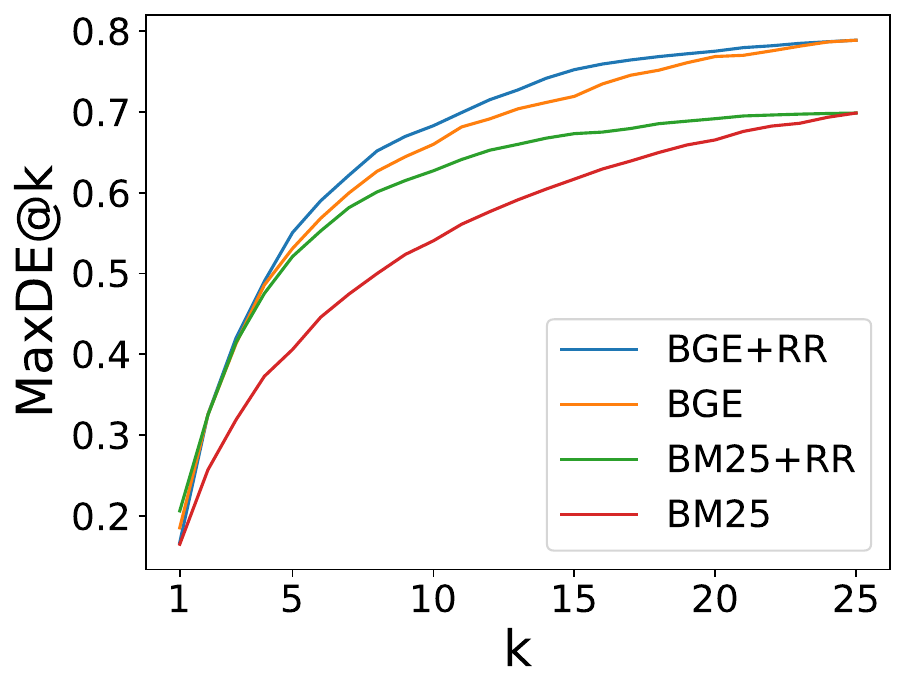}}
    \subfigure[]{\includegraphics[width=0.24\textwidth]{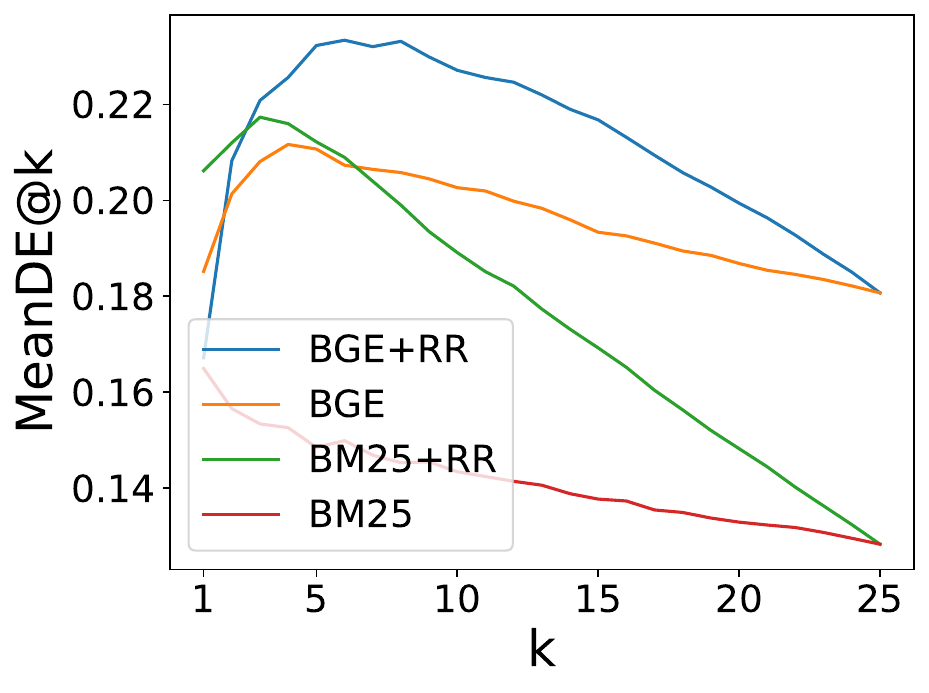}}

    \caption{Results of different retrieval pipelines when varying the number $k$ of retrieved passages. We compute the distracting effect on Qwen 2.5 7B.} %Results are averaged across PopQA, TriviaQA, and NQ. Results on each specific dataset are available in the Appendix \ref{sec:add_retr_res}. }
    \label{fig:retrieval_performance}
\end{figure*} 
\section{Related work}
% \begin{itemize}
%     \item Report literature on distracting effect, including lost-in-the-middle and many works that attempt to mitigate this bias. Say the actual limits of these works, e.g., outdated LLM where the bias was more pronounced (Is it true?), non-realistic evaluation (i.e., one relevant and X random, or weird dataset - key-value pairs), analysis not considering the presence of distracting documents.
%     \item report literature on distracting effect e.g., \cite{cuconasu2024power} and our ACL paper.
% \end{itemize}

\paragraph{Effect of Irrelevant Content.} Recent work explores the detrimental effect of irrelevant content in the LLM prompt. In the RAG setting, a passage is considered irrelevant if it does not provide useful information for answering the query. \citet{cuconasu2024power} divide irrelevant passages as either random, if they are semantically unrelated to the query, or distracting, if they are related to the query but do not contain the answer. They show that while random passages do not affect answer quality, distracting passages do. \citet{jin2025longcontext} show that irrelevant passages returned by strong retrievers are more detrimental than those obtained by weak retrievers. 
\citet{amiraz2025distracting} propose a continuous measure of the distracting effect of irrelevant passages and a fine-tuning approach to enhance LLM robustness, similar to strategies in \cite{linra2024, jin2025longcontext, yoran2024making}.
To mitigate these challenges, several approaches have emerged to compress or filter retrieved content: \citet{yu-etal-2024-chain} generate sequential reading notes that evaluate document relevance before answer generation, \citet{xu2024recomp} compress retrieved documents into concise textual summaries using both extractive and abstractive methods with selective augmentation, and \citet{huang2024fastfid} perform sentence-level selection from encoded passages to reduce context length while preserving inference quality.

\paragraph{Positional Bias.} Despite advanced positional encoding methods like Alibi \cite{press2022train} and RoPE \cite{10.1016/j.neucom.2023.127063}, long-context LLMs are typically affected by position bias, i.e., their capability of identifying relevant content depends on its location in the prompt. \citet{liu-etal-2024-lost} discuss the \textit{lost-in-the-middle} effect, where the LLMs tend to ignore information in the middle of the prompt. \citet{hutter-2025-lost-but-not-only} extend this work and demonstrate that different LLMs exhibit distinct positional bias patterns.
% , including patterns that differ from the lost-in-the-middle U-shape.  
To mitigate this bias, some solutions propose to fine-tune the LLMs on training data where relevant information is equally distributed across all positions of the prompt \cite{he-etal-2024-never, an2024make}. Other methods modify the attention mechanism of the transformer architecture to remove token-level bias \cite{leviathan2025selective, ye2025differential}. \citet{peysakhovich2023attentionsortingcombatsrecency} propose a double decoding approach, where in the second decoding step, the passages are re-ordered based on the attention they received in the first step. \citet{jin2025longcontext} re-order the retrieved passages so that top-ranked passages are placed in privileged positions according to the lost-in-the-middle behavior. \citet{zhang-etal-2024-instruct} instruct the LLM directly in the prompt to allocate more attention towards a selected segment of the context, aiming to compensate for the shortage of attention. \citet{jiang-etal-2024-longllmlingua} mitigates the positional bias by introducing an external module to compress the prompt.

\section{Experimental Setup} \label{sec:setup}

\paragraph{Benchmarks and Models.} 
We run experiments using the following commonly used public question-answering benchmarks: PopQA \cite{mallen2023not} and the KILT version \cite{petroni-etal-2021-kilt} of Natural Questions (NQ) \cite{kwiatkowski2019natural}, and TriviaQA \cite{joshi2017triviaqa}. From each benchmark, we randomly select two disjoint 500-size samples to run the experiments in Sections \ref{sec:controlled} and \ref{sec:real}, respectively. The results we report in the main paper are averaged across the three datasets\footnote{Appendix \ref{sec:add_retr_res} provides results on each benchmark.}. 
We index the corpus\footnote{Further details about corpus processing in Appendix \ref{sec:corpus}.} using BM25 \cite{bm25} for sparse retrieval and the \emph{BGE large en v1.5} embedding model \cite{chen2024bge} for dense retrieval.
% We index the corpus\footnote{Further details about corpus processing in Appendix \ref{sec:corpus}.} using both sparse and dense retrieval approaches. We use BM25 for sparse retrieval, and the \emph{BGE large en v1.5} embedding model \cite{chen2024bge} for dense retrieval. % with  Opensearch\footnote{\url{https://opensearch.org}}
%while for dense retrieval, we use . % and the Pinecone vector DB\footnote{\url{https://www.pinecone.io/}}
Additionally, we used a re-ranker (RR), namely \emph{BGE reranker v2 m3} \cite{chen2024bge}, to rerank the first 25 results from the retriever.

We estimate the performance of the four retrieval pipelines in terms of HITS@$k$ in Fig.~\ref{fig:retrieval_performance}a, measuring the percentage of times at least a relevant passage is in the top-$k$ retrieved ones, and Precision@$k$ in Fig.~\ref{fig:retrieval_performance}b, measuring the average percentage of relevant passages in the top-$k$ retrieved ones. Especially when the re-ranker is used, HITS plateaus soon, while Precision keeps decreasing since low-ranked passages are mostly irrelevant. This suggests that using large values of $k$ (e.g., beyond 10) is not worth it, as this would simply add irrelevant passages to the prompt.
% Therefore, the experimental evaluations in the next sections will focus on two reasonable values for $k$, namely 5 and 10, which provide a good compromise for practical RAG applications in terms of accuracy and latency.
Therefore, our experiments focus on two reasonable values for $k$, namely 5 and 10, which provide a good accuracy-latency tradeoff.

% As LLMs, we conduct experiments with the instruct version of Llama 3.2 3B (L3B), Llama 3.1 8B (L8B), Llama 3.3 70B (L70B) \cite{grattafiori2024llama3herdmodels}, and Qwen 2.5 7B (Q7B) \cite{qwen2025qwen25technicalreport}. This selection includes LLMs of different sizes (from 3B to 70B), and family (Llama, and Qwen), enabling a comprehensive evaluation. 
As LLMs, we use the instruction-tuned version of Llama 3.2 3B (L3B), Llama 3.1 8B (L8B), Llama 3.3 70B (L70B) \cite{grattafiori2024llama3herdmodels}, and Qwen 2.5 7B (Q7B) \cite{qwen2025qwen25technicalreport}, spanning different model sizes and families.

% \begin{figure}[t!]
%     \centering
%     \includegraphics[width=\linewidth]{img/qwen_2.5_7B_relevant.pdf}
%     \caption{Qwen 2.5 7B averaged across datasets}
%     \label{fig:relevant_and_hard_in_weak}
% \end{figure}

\begin{figure}[t!]
    \centering
    \subfigure[]{
    \includegraphics[width=0.48\linewidth]{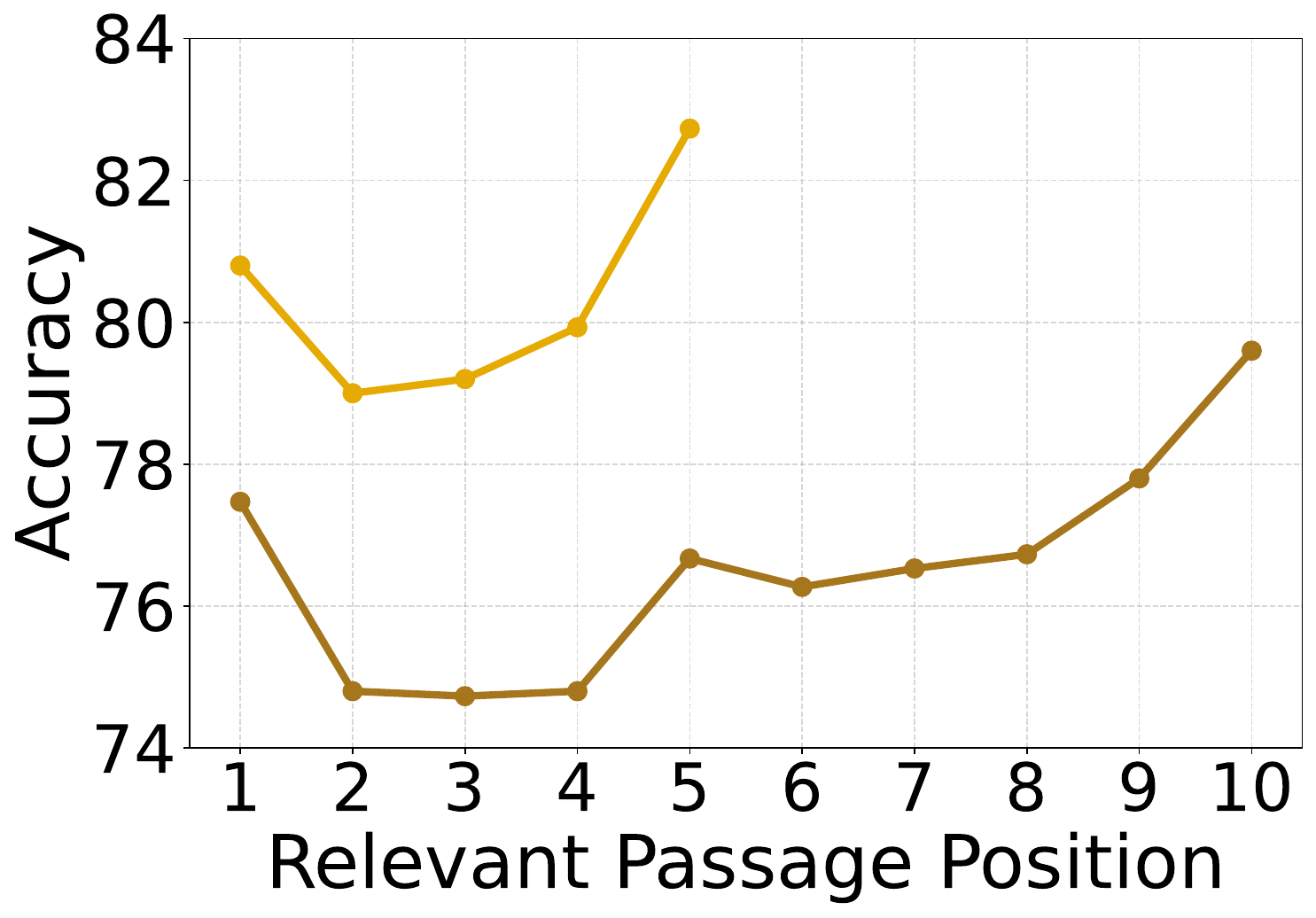}}
    \subfigure[]{
    \includegraphics[width=0.48\linewidth]{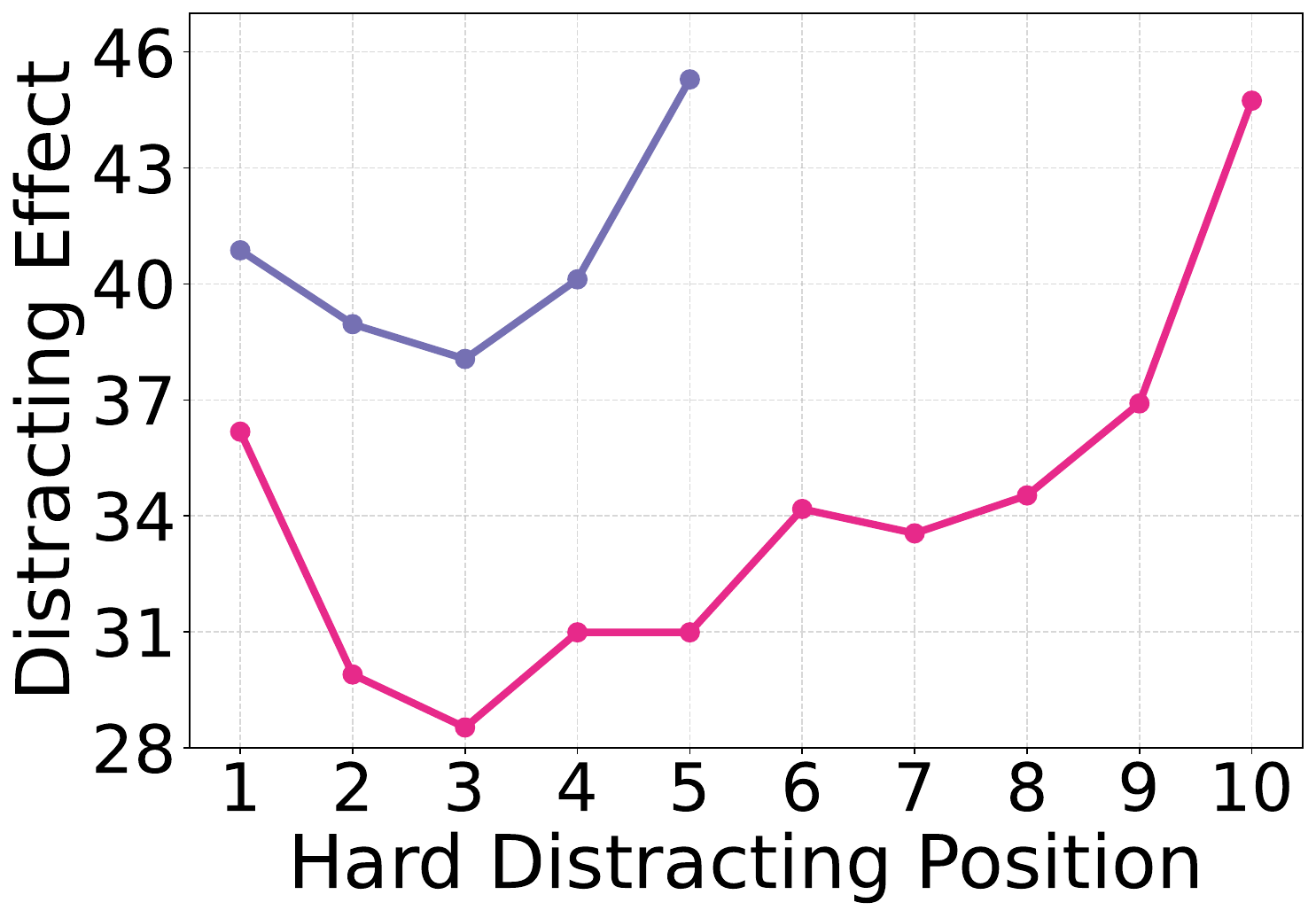}}

    \caption{Controlled experiments results for Qwen 2.5 7B. \textbf{(a)} Average accuracy when rotating a single relevant passage among weak distractors. \textbf{(b)} Average distracting effect when rotating a hard distractor among weak distractors. Both exhibit the characteristic U-shaped positional bias pattern.}
    \label{fig:relevant_and_hard_in_weak}
\end{figure}

\paragraph{Evaluation Strategy.} 
Following related work \cite{zheng2023judging,gu2025surveyllmasajudge,rahmani2024report1stworkshoplarge}, we evaluate passage relevance and answer quality using the LLM-as-a-judge approach. In the former case, we prompt the LLM to assess the relevance of a passage to a question given the ground truth answer, where, following \citet{cuconasu2024power}, we consider a passage relevant if and only if it contains the answer to the question, and irrelevant otherwise. In the latter, we prompt the LLM to assess whether the generated response semantically matches the reference answer\footnote{Exact prompts are provided in Appendix \ref{sec:llm_as_judge}.}. We use Claude 3.7 Sonnet via AWS Bedrock as the backbone LLM.\footnote{We use Claude 3.7 Sonnet to minimize evaluation errors, though strong open-source models like Llama 3.3 70B Instruct achieve comparable performance (see Appendix \ref{sec:llm_judge_comparison}).}

During the experiments, we use the definition of distracting effect introduced by \citet{amiraz2025distracting}. Specifically, their approach consists of prompting an LLM to answer a question $q$ using the information from a passage $p$ or abstain (output ``NO-RESPONSE'') if the passage does not contain an answer to $q$. The distracting effect $\text{DE}_q(p)$ of an \emph{irrelevant} passage $p$ for question $q$ is then computed as the probability of the LLM not abstaining:
\vspace{-0.1cm}
\begin{equation}
\label{eq:distracting_effect}
    \text{DE}_q(p) = 1 - p^{\text{LLM}}(\text{NO-RESPONSE}|q,p)
\vspace{-0.1cm}
\end{equation}
For each retrieval pipeline, we compute the distracting effect only of irrelevant passages, and assign DE=0 for relevant passages.
% For each retrieval pipeline, we compute the distracting effect of the retrieved irrelevant passages and assume DE=0 for relevant passages.
Fig. \ref{fig:retrieval_performance}c reports the DE of the most distracting passage among the top-$k$ positions (MaxDE), while Fig. \ref{fig:retrieval_performance}d reports the mean DE considering the top-$k$ positions (MeanDE). Both metrics are averaged across all queries. The MaxDE curves reach very high values, 
% showing that for most of the queries, at least a very distracting passage exists among the retrieved ones. 
with Table \ref{tab:hard_distractor_prevalence} (Appendix) showing that over 60\% of queries contain at least one hard distractor (defined as having a DE score greater than 0.8) in the top-10 results from dense retrievers. The MeanDE curves are initially very low, since most of the top retrieved passages are relevant, then increase as more irrelevant passages appear in the prompt, but soon they decrease again. This suggests that highly distracting passages typically appear in top positions, while low-ranked passages have a DE score close to 0. Finally, retrieval pipelines leading to higher HITS and Precision, e.g., when using BGE, also exhibit higher MaxDE and MeanDE curves, revealing a critical aspect: \textit{stronger retrievers increase recall \textbf{and} deliver more harmful distractors, making retrieval a double-edged sword}.

\begin{figure*}[t!]
\begin{mdframed}[font=\small]
\texttt{You are given a question and you must respond based on the provided documents. Respond directly without providing any premise or explanation.\\
\\Documents:}

\vspace{0.2cm}
\begin{tabular}{@{}p{1.2cm}@{\hspace{0.10cm}}p{14cm}@{}}
\small DE=0.13 & \texttt{\textbf{Document[1]} (Title: Bids for the 2024 and 2028 Summer Olympics)(Section: Non-selected bids - 2024 - United States) [...] Following the final presentation, the USOC announced that the United States would bid to host the 2024 Olympic and Paralympic Games, but did not announce which city would bid. On 8 January 2015, the USOC selected Boston to be the candidate city from the United States but on 27 July 2015 Boston's bid was withdrawn and the USOC bid process was reopened. On 1 September 2015 the USOC announced that Los Angeles was chosen for the United States bid for the 2024 Summer Games.} \\[0.3cm]

\footnotesize \textcolor{gold}{DE=0.00} & \texttt{\textbf{\textcolor{gold}{Document[2]}} (Title: Sports in the United States)(Section: Olympics) [...] The United States hosted both Summer and Winter Games in 1932, and has hosted more Games than any other country – eight times, four times each for the Summer and Winter Games: BULLET::::- the 1904 Summer Olympics in St. Louis, 1932 Summer Olympics and 1984 Summer Olympics in Los Angeles; and the 1996 Summer Olympics in Atlanta; BULLET::::- the 1932 Winter Olympics and 1980 Winter Olympics in Lake Placid, New York; the 1960 Winter Olympics in Squaw Valley, California; and the \textcolor{gold}{2002} Winter Olympics in Salt Lake City, Utah. Los Angeles will host the Summer Olympics for a third time in 2028, marking the ninth time the U.S. hosts the Olympic Games.} \\[0.3cm]

\footnotesize DE=0.01 & \texttt{\textbf{Document[3]} (Title: 1992 Winter Olympics)(Section: Legacy) The 1992 Olympic Winter Games marked the last time both the Winter and Summer games were held in the same year. The 1992 Olympics also marks the last time France hosted the Olympics. Paris will host the 2024 Summer Olympics.} \\[0.3cm]

\footnotesize DE=0.19 & \texttt{\textbf{Document[4]} (Title: Sports in Chicago)(Section: Olympic bids) [...] Following Chicago's loss in the race for the 2016 Olympics, the USOC bid for the 2024 Olympics with Los Angeles which result in a deal where Los Angeles secured the right to host the 2028 Summer Olympics. Chicago had previously hosted the 1959 Pan American Games. Chicago was selected to host the 1904 Summer Olympics, but they were transferred to St. Louis to coincide with the Louisiana Purchase Exposition.} \\[0.3cm]

\footnotesize \textcolor{distractor}{DE=0.98} & \texttt{\textbf{\textcolor{distractor}{Document[5]}} (Title: Summer Olympic Games)(Section: Hosting) The United States has hosted the Summer Olympic Games four times: the 1904 Games were held in St. Louis, Missouri; the 1932 and 1984 Games were both held in Los Angeles, California; and the \textcolor{red}{1996} Games were held in Atlanta, Georgia. The 2028 Games in Los Angeles will mark the fifth occasion on which the Summer Games have been hosted by the U.S. [...]} \\[0.3cm]
\end{tabular}

\vspace{0.2cm}
\texttt{\textbf{Question}: When did the united states host the last olympics?\\
\textbf{Answer}: The United States hosted the last Summer Olympics in \textcolor{red}{1996} in Atlanta, Georgia.\\ \\
\textbf{Gold Answer}: \textcolor{gold}{2002}}
\end{mdframed}
\caption{Example showing how the position of the hard distractor affects Qwen 2.5 7B's response when a relevant passage is fixed in position 2. When a hard distractor (Document 5, DE=0.98) is placed in position 5 (highest distracting effect according to Fig. \ref{fig:relevant_and_hard_in_weak}b), the model provides an incorrect answer based on the hard distractor. However, simply moving the hard distractor to position 3 (lowest distracting effect), while maintaining the relevant passage in position 2, results in the model correctly answering ``2002''.}
\label{fig:main_example_prompt}
\end{figure*}

\section{Positional Bias in Controlled Settings} \label{sec:controlled}

\begin{table}[t]
\centering
\resizebox{\columnwidth}{!}{
\begin{tabular}{cccccc}
\toprule
\multirow{2}{*}{\textbf{Hard Distractor}} & \multicolumn{5}{c}{\textbf{Relevant Passage Position}} \\
\cmidrule{2-6}
& \textbf{1} & \textbf{2} & \textbf{3} & \textbf{4} & \textbf{5} \\
\midrule
\textbf{None} & 80.80 & 79.00 & 79.20 & 79.93 & 82.73 \\
\textbf{Position 3} & 75.13 & 73.80 & - & 72.40 & 76.73 \\
\textbf{Position 5} & 72.87 & 71.53 & 71.60 & 73.20 & - \\
\bottomrule
\end{tabular}
}
\caption{Answer accuracy of Qwen 2.5 7B when rotating a relevant passage in weak distractors only (None), and in weak distractors and a single hard distractor at position 3 or 5.}
\label{tab:both_relevant_and_hard_in_weak}
\end{table}

While previous work has established the existence of positional bias in LLMs \cite{liu-etal-2024-lost, hsieh2024found, hutter-2025-lost-but-not-only}, these studies typically only analyze the problem from the viewpoint of the relevant passages and completely neglect how the positional bias impacts the effect of distracting passages. In this work, we present the first systematic investigation of the impact of positional bias on distracting passages, analyzing their interactions with relevant content.
%examine how the change in position of the relevant document affects answer accuracy. We argue that the performance is affected not only by where relevant information appears but also by the positioning of highly distracting documents.
% \simo{below your first provide a high-level intro of the two experiments, and then you present them in detail. I don't think we have space for this, consider cutting the high-level intro.}

For each query, we select the highest-ranked relevant passage obtained by BGE after reranking (BGE+RR). Following \citet{amiraz2025distracting}, we compute the distracting effect for irrelevant passages using Equation \ref{eq:distracting_effect}. We classify passages as ``hard distractors'' (with DE $> 0.8$, as previously defined) and ``weak distractors'' (with DE $< 0.2$). As an example, Fig. \ref{fig:main_example_prompt} illustrates this distinction: while weak distractors (Documents 1, 3, 4 with DE=0.01-0.19) have minimal impact on the model's reasoning, hard distractors (Document 5 with DE=0.98) can cause the LLM to overlook correct information from relevant passages and generate incorrect answers.
Fig. \ref{fig:relevant_and_hard_in_weak} shows results for Qwen 2.5 7B (results for other models and single datasets are given in Appendix \ref{sec:results_other_llms}). Fig. \ref{fig:relevant_and_hard_in_weak}a displays the characteristic U-shaped accuracy pattern when rotating a single relevant passage among fixed weak distractors\footnote{We use weak distractors instead of general retrieved irrelevant passages to avoid negative effects from hard distractors.}.
% \footnote{We use weak distractors instead of general retrieved irrelevant passages to avoid the potential negative influence of hard distractors.}
Fig. \ref{fig:relevant_and_hard_in_weak}b shows that this positional bias extends to distracting passages, with hard distractors at the beginning or end having significantly higher distracting effect (36-44\%) compared to middle slots (28-34\%)\footnote{We calculate the distracting effect using Equation \ref{eq:distracting_effect} applied to the entire set of passages rather than a single passage.}.
% \footnote{To compute the distracting effect, we use the same logic behind Equation \ref{eq:distracting_effect}, but instead of computing the probability of abstention using a single passage, we calculate it over the entire set of passages, i.e., 5 or 10 in our experiments.}.
% This parallel pattern suggests the model's attention mechanism gives preferential treatment to information at certain positions regardless of relevance quality, and shows that the damaging effect of hard distracting passages can be amplified based on their position.
This parallel pattern indicates the model favors certain positions regardless of passage relevance.

Table \ref{tab:both_relevant_and_hard_in_weak} further validates this point by showing accuracy when placing a hard distractor at position 3 (lowest DE) versus position 5 (highest DE). We observe an average decrease of about 6 accuracy points compared to using only weak distractors (first row of the table), with a more pronounced drop when the hard distractor occupies position 5. This confirms how positional preference amplifies the negative impact of distracting content.

\section{Positional Bias in Real Scenarios} \label{sec:real}

% \begin{table}[t]
% \setlength{\tabcolsep}{3.2pt}
% \footnotesize
% \begin{tabular}{lccccc}
% \textbf{LLM}                    & \textbf{Sequential}  & \textbf{Inverse}   & \textbf{Shuffle } & \textbf{MaxRel} & \textbf{MinDist} \\
% \hline
% Q7B   & 68.53 & 71.33 & 71.00 & 71.73 & 70.80                \\
% L3B   & 65.80 & 68.00 & 66.73 & 67.33 & 66.20               \\
% L8B  & 69.13 & 69.60 & 69.87 & 69.60 & 69.27                \\
% L70B  & 74.33 & 74.40 & 74.60 & 74.33 & 75.47    
% \end{tabular}

% \caption{Answer accuracy when arranging the top-5 passages retrieved by BGE+RR using different strategies. 
% % Results with $k$=10 are available in the Appendix \ref{sec:app_real_setting}.
% }
% \label{tab:all_llms_res}
% \end{table}

\begin{table}[t]
\centering
\setlength{\tabcolsep}{3.2pt}
\footnotesize
\begin{tabular}{@{}lccccc@{}}
\toprule
\textbf{LLM} & \textbf{Sequential} & \textbf{Inverse} & \textbf{Shuffle} & \textbf{MaxRel} & \textbf{MinDist} \\ \midrule
Q7B  & 68.53 & 71.33 & 71.00 & 71.73 & 70.80 \\
L3B  & 65.80 & 68.00 & 66.73 & 67.33 & 66.20 \\
L8B  & 69.13 & 69.60 & 69.87 & 69.60 & 69.27 \\
L70B & 74.33 & 74.40 & 74.60 & 74.33 & 75.47 \\ \bottomrule
\end{tabular}
\caption{Answer accuracy when arranging the top-5 passages retrieved by BGE+RR using different strategies.}
\label{tab:all_llms_res}
\end{table}

% Attempt to provide a formula for computing the expected accuracy of a system, given the positional bias and the retrieval quality. Limit this analysis to the cases in which there is at least a relevant passage. Show how positioning in the prompt is expected to affect results and the differences that are expected to occur between different positioning strategies (for instance, emphasize what the expected margin is between the vanilla positioning and the position-bias driven positioning). Run real experiments with different ordering strategies. Say that the expected differences, which are already not huge, do not really occur in practice because of (i) the distracting effect and (ii) the possibility that multiple relevant docs appear in the prompt.

% TODO: we need results on multiple datasets (e.g., 3). It would be nice if one dataset is not QA.

% In this section, we want to address the following research question: \emph{What is the practical impact of the LLM's positional bias in a RAG setting?}

In Section \ref{sec:controlled}, we showed how the answer accuracy can vary up to 5 percentage points in controlled settings, depending on the relevant passage's position. Here, instead, we study the impact of position in real RAG scenarios, i.e., when the LLM prompt contains the top-$k$ ranked passages from the retrieval pipeline. This setting is substantially different from the controlled one shown in Fig. \ref{fig:relevant_and_hard_in_weak}a. Indeed, 
%ym rephrased as it did to parse well, review to validate?
%\Ni there is no guarantee to have a unique relevant passage, but there can be 0 or multiple ones; \Nii there can be one or more highly distracting passages confusing the LLM. 
there is no guarantee that a single relevant passage occurs among the top-$k$ ranked passages: there could be none or multiple ones, as well as one or more highly distracting passages.
Therefore, we arrange the top-$k$ retrieved passages in the LLM prompt according to the following strategies: (\textit{i}) \texttt{Shuffle}: random ordering of passages;
% the top-$k$ retrieved passages are placed in random order.
(\textit{ii}) \texttt{Sequential}: maintaining retrieval ranking order;
% the retrieval order is used ``as is''.
(\textit{iii}) \texttt{Inverse}: inverting the retrieval order, so that according to our LLM prompt template (Fig. \ref{fig:prompt dist test}), the top-1 retrieved passage is the closest to the question; (\textit{iv}) \texttt{MaxRelevance}: ranking passages by decreasing positional accuracy estimated during the controlled experiments with the relevant passage\footnote{For example, following Fig. \ref{fig:relevant_and_hard_in_weak}a for Qwen 2.5 7B with $k=5$, the estimated order would be \textit{5, 1, 4, 3, 2}.}.
% the retrieved passages are ranked by decreasing order of positional accuracy estimated during the controlled experiments about rotating the relevant passage
%\ym{not clear what is estimated here, the accuracy?}\simo{We estimated the accuracy associated with each position when it contains the relevant doc, and based on these accuracy scores, we estimate the optimal order to follow when arranging the passages.} estimated in the controlled setting.
%In other words, we learn the preferred positions for the relevant passage, and, at inference time, we organize the top-$k$ retrieved passages to follow this order\footnote{As an example, following Fig. \ref{fig:relevant_and_hard_in_weak}a for Qwen 2.5 7B with $k=5$ the estimated order would be \textit{5, 1, 4, 3, 2}.}. 
Assuming the retrieval pipeline ranks the passages by decreasing probability of relevance, this strategy maximizes the likelihood of having relevant passages in LLM-favored slots;
%ym @simone replaced favoured with favored everywhere as we use US and not UK spelling:) 
\Nv \texttt{MinDistraction}: arranging passages by increasing DE order estimated in the controlled setting\footnote{As an example, following Fig. \ref{fig:relevant_and_hard_in_weak}b for Qwen 2.5 7B with $k=5$ the estimated order would be \textit{3, 2, 4, 1, 5}.}. Assuming that the retrieval pipeline ranks passages by decreasing DE (as evident in Fig.~\ref{fig:retrieval_performance}d), this strategy minimizes the likelihood of having highly distracting passages in LLM-favored positions.

% \begin{table}[t]
% \setlength{\tabcolsep}{1.6pt}
% \footnotesize
% \begin{tabular}{lccccc}
% \textbf{Retriever}                  & \textbf{Sequential}  & \textbf{Inverse}   & \textbf{Shuffle } & \textbf{MaxRel} & \textbf{MinDist} \\
% \hline
% BGE & 68.00	& 69.00	& 68.40	& 68.80 & 67.47 \\
% BGE+RR & 68.53 & 71.33 & 71.00 & 71.73 & 70.80                \\
% BM25 & 51.20 &	51.27 &	51.00 &	 51.00 & 51.00 \\
% BM25+RR & 59.27 & 60.20	& 59.80 & 59.80 & 58.80 \\
             
% \end{tabular}

% \caption{Answer accuracy of Qwen 2.5 7B when arranging with different strategies the top-5 passages retrieved from different retrieval pipelines. 
% % Results with $k$=10 are available in the Appendix \ref{sec:app_real_setting}.
% }
% \label{tab:all_pipelines_res}
% \end{table}

\begin{table}[t]
\centering
\setlength{\tabcolsep}{1.6pt}
\footnotesize
\begin{tabular}{@{}lccccc@{}}
\toprule
\textbf{Retriever} & \textbf{Sequential} & \textbf{Inverse} & \textbf{Shuffle} & \textbf{MaxRel} & \textbf{MinDist} \\ \midrule
BGE     & 68.00 & 69.00 & 68.40 & 68.80 & 67.47 \\
BGE+RR  & 68.53 & 71.33 & 71.00 & 71.73 & 70.80 \\
BM25    & 51.20 & 51.27 & 51.00 & 51.00 & 51.00 \\
BM25+RR & 59.27 & 60.20 & 59.80 & 59.80 & 58.80 \\ \bottomrule
\end{tabular}
\caption{Answer accuracy of Qwen 2.5 7B when arranging with different strategies the top-5 passages retrieved from different retrieval pipelines.}
\label{tab:all_pipelines_res}
\end{table}

Results in Tables \ref{tab:all_llms_res} and \ref{tab:all_pipelines_res} show that the impact of the positional bias in real settings is minor: different passage arrangement strategies lead to very similar results that do not significantly differ from the \texttt{Shuffle} baseline\footnote{Statistical significance using Wilcoxon test with $p$=0.05.}, regardless of the LLM or the retrieval pipeline. We argue that these results can be explained by the contrastive effect of relevant and highly distracting passages, which, as observed in Fig. \ref{fig:retrieval_performance}, tend to both appear in top retrieved passages: for instance, in the \texttt{MaxRelevance} strategy, the benefit of placing relevant passages in LLM-favored positions is compensated by the unintended tendency to put in the same slots highly distracting passages.

\section{Conclusions} \label{sec:conclusions}
Our work demonstrates that while positional bias exists in current LLMs, its impact is minimal in realistic RAG settings: random ordering of retrieved passages yields statistically equivalent accuracy to more sophisticated reordering strategies.
%We showed that the mean distracting effect is substantially greater for top-ranked passages than for those in lower ranks.
%ym removed as we say it again later
%turning retrieval into a potential source of error rather than a helper. 
We observed that contemporary
%sparse, dense, and reranked 
retrievers do not merely return some irrelevant passages, they 
%\textit{systematically} 
surface passages that degrade answer accuracy in more than 60\% of our test questions, turning the retriever itself into a first‑order source of error. 
Thus, attempting to place relevant passages in LLMs' favorable positions may inadvertently prioritize hard distractors over relevant content, counterbalancing the potential benefits of strategic reordering. 
%ym removing as not accurate as per simone
%In up to 80\% of the questions we trested, at least one of the top-5 passages actively reduces the mode's likelihood of abstaining,  elevating the retriever itself to a first‑order source of error. 
%Thus, the interaction between relevant and distracting passages counterbalances potential benefits from strategic reordering. Placing passages in LLMs' favorable positions may inadvertently prioritize hard distractors over relevant content due to current retrieval pipeline limitations. 
These findings suggest that future improvements should focus on retrieval quality and LLM distraction robustness rather than passage positioning.

\clearpage
\newpage
\section*{Limitations}

Our research primarily investigates the factoid question-answering task, though the concept of distracting passages applies to various RAG use cases. Indeed, extending the study to additional tasks, such as multi-hop question answering or fact verification, will provide a more complete picture, but we defer that to future work. Additionally, while we conducted our experiments on English-language benchmarks, the language-agnostic nature of our methodology suggests that the findings would likely generalize to other languages, though formal verification of this hypothesis remains as future work.

\section*{Acknowledgments}
This work was carried out while Florin Cuconasu was enrolled in the Italian National Doctorate on Artificial Intelligence run by the Sapienza University of Rome. This project has also been supported by PNRR MUR project PE0000013-FAIR and PE00000014-SERICS, funded by European Union – NextGenerationEU, and the NEREO PRIN project funded by the Italian Ministry of Education and Research Grant no. 2022AEFHAZ.

% Bibliography entries for the entire Anthology, followed by custom entries
%\bibliography{anthology,custom}
% Custom bibliography entries only
\bibliography{custom}

\appendix

\section{Additional Details on the RAG Pipeline} \label{sec:appendix}

\subsection{Corpus and Chunking} \label{sec:corpus}
We use the KILT knowledge base\footnote{\href{https://huggingface.co/datasets/facebook/kilt_wikipedia}{https://huggingface.co/datasets/facebook/kilt\_wikipedia}} as corpus for our retrieval. It corresponds to the Wikipedia dump of 01 August 2019. It comprises 5,874,358 Wikipedia articles, which we chunk using SentenceSplitter by LlamaIndex\footnote{\href{https://docs.llamaindex.ai/en/stable/api_reference/node_parsers/sentence_splitter/}{https://docs.llamaindex.ai/en/stable/api\_reference/\\node\_parsers/sentence\_splitter/}} with a chunking size of 256 and no overlap. The splitter tries to segment chunks based on full sentences, avoiding truncations in the middle of a phrase. 
The chunking phase produced 27,492,989 passages. Then, we index the corpus using Opensearch\footnote{\href{https://opensearch.org}{https://opensearch.org}} for sparse retrieval and Pinecone\footnote{\href{https://www.pinecone.io/}{https://www.pinecone.io/}} for dense retrieval. 

When prompting an LLM with a retrieved passage, we augment it with the title and subsection names from Wikipedia to provide more contextual information to each individual segment (see an example in Fig. \ref{fig:main_example_prompt}).

\subsection{Additional Retrieval Results}\label{sec:add_retr_res} 
\begin{figure*}
    \centering
    \subfigure[]{\includegraphics[width=0.24\textwidth]{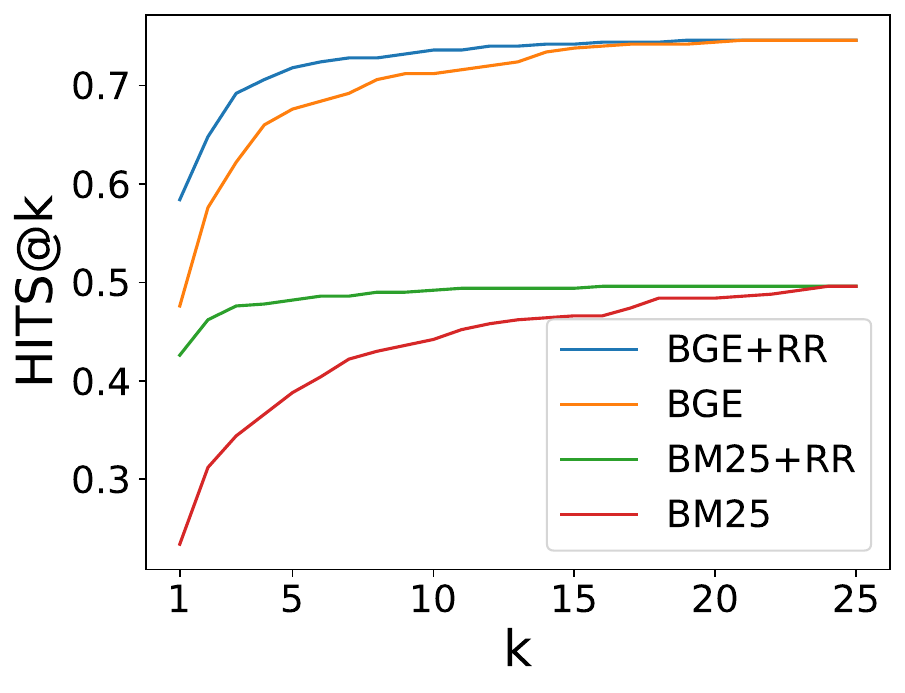}} 
    \subfigure[]{\includegraphics[width=0.24\textwidth]{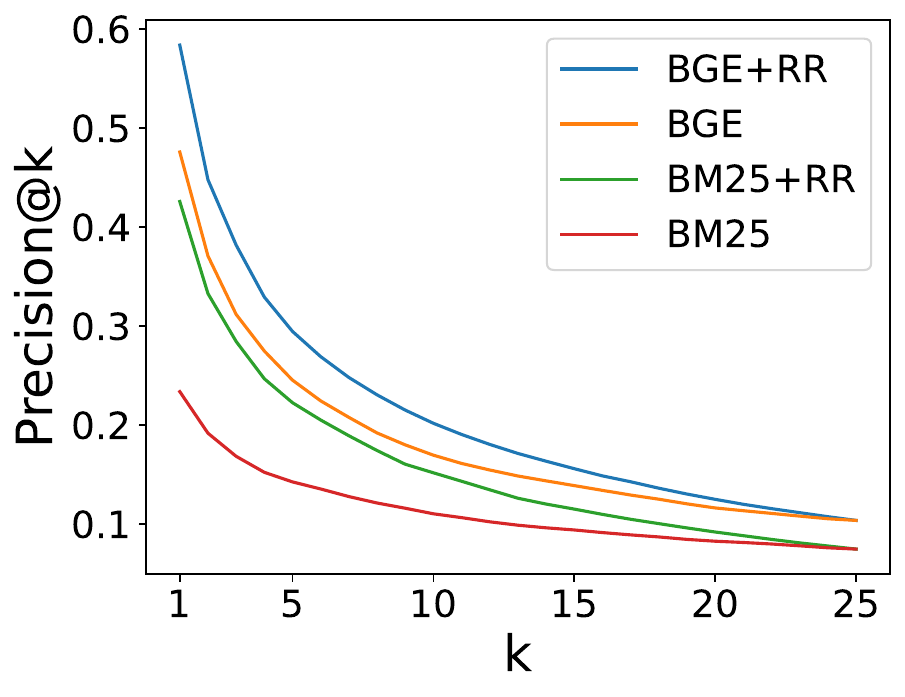}} 
    \subfigure[]{\includegraphics[width=0.24\textwidth]{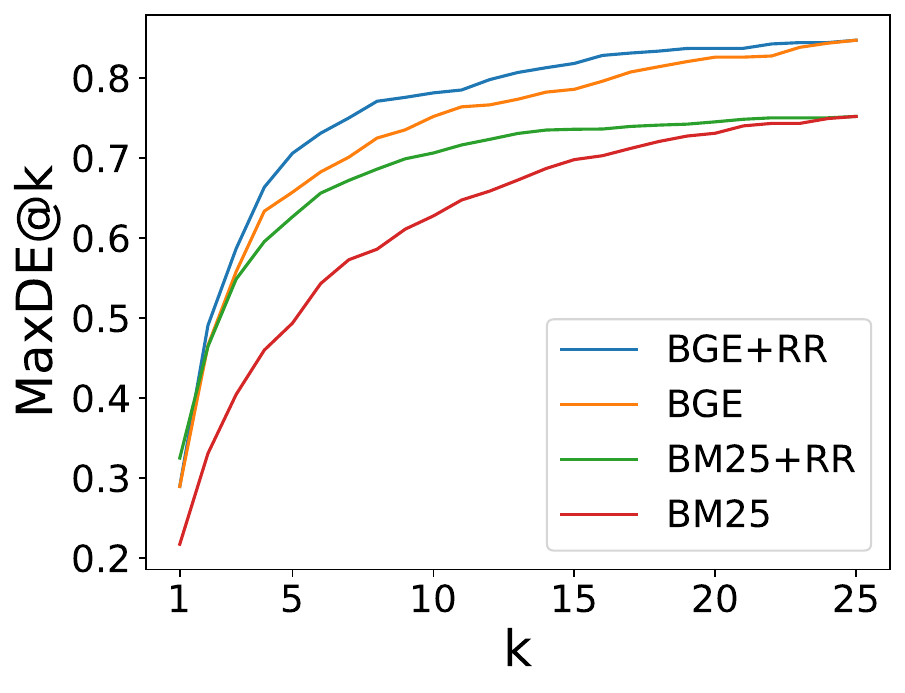}}
    \subfigure[]{\includegraphics[width=0.24\textwidth]{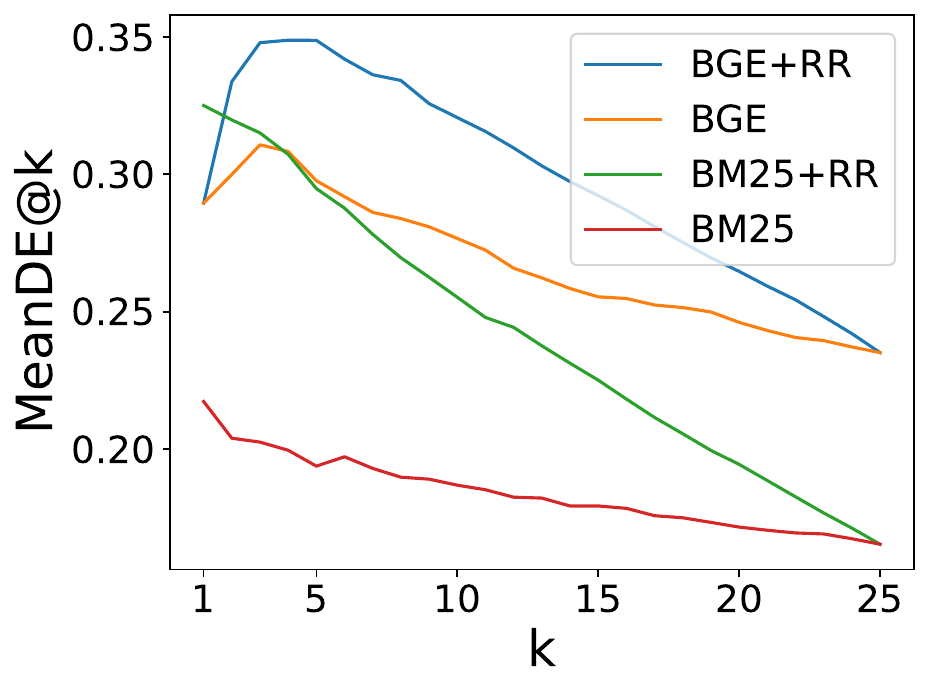}}
    \caption{Results on PopQA of different retrieval pipelines when varying the number $k$ of retrieved passages. We compute the distracting effect on Qwen 2.5 7B.}
    \label{fig:retrieval_performance_on_POPQA}
\end{figure*}

\begin{figure*}
    \centering
    \subfigure[]{\includegraphics[width=0.24\textwidth]{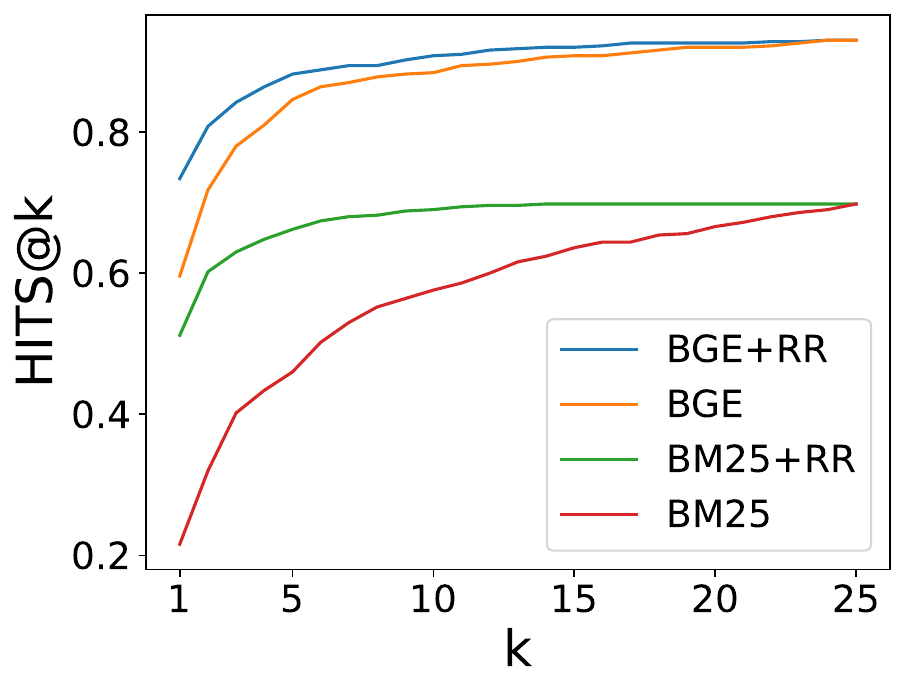}} 
    \subfigure[]{\includegraphics[width=0.24\textwidth]{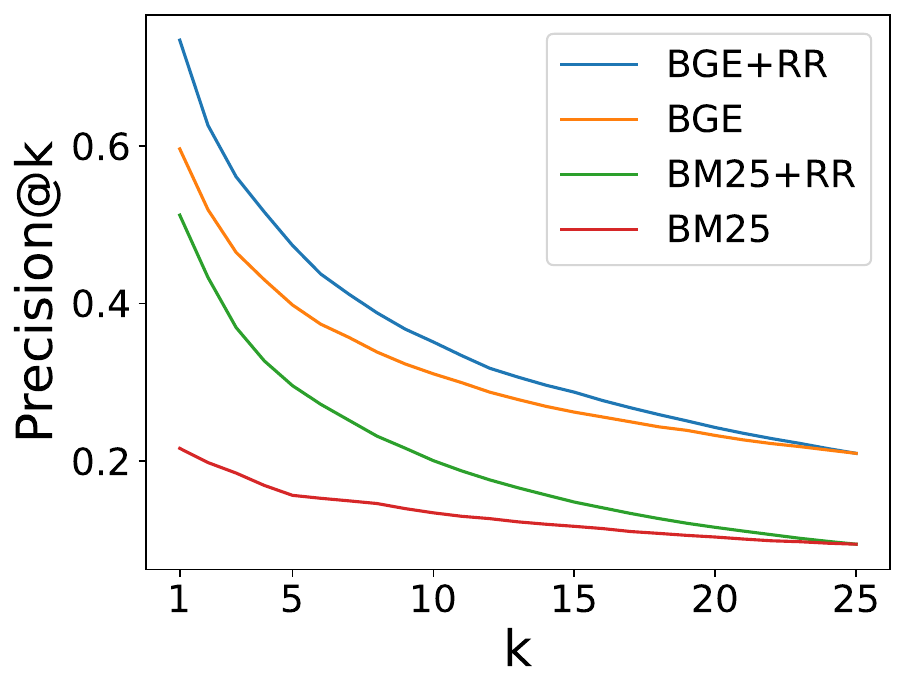}} 
    \subfigure[]{\includegraphics[width=0.24\textwidth]{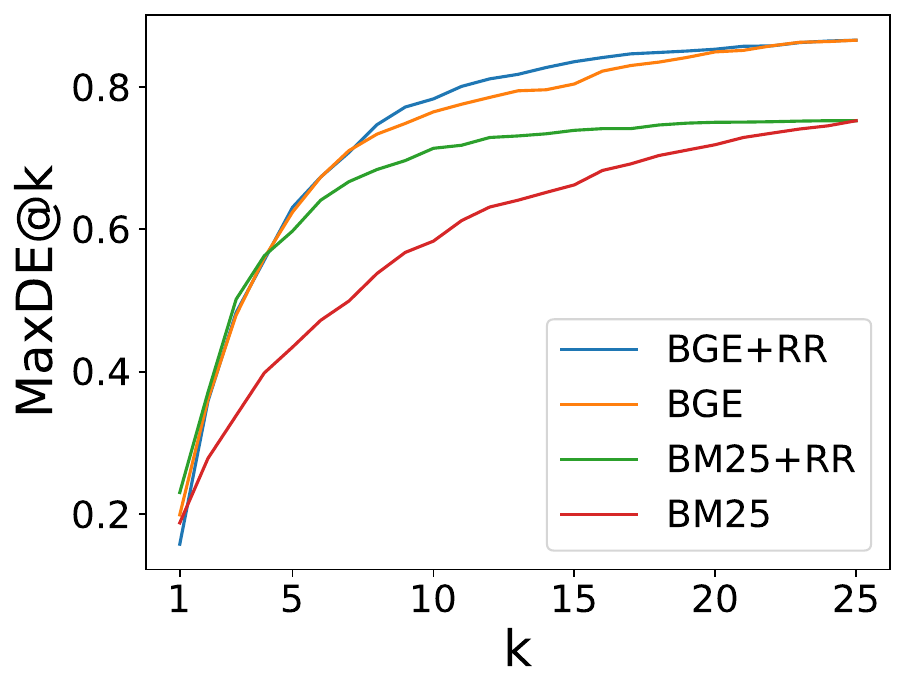}}
    \subfigure[]{\includegraphics[width=0.24\textwidth]{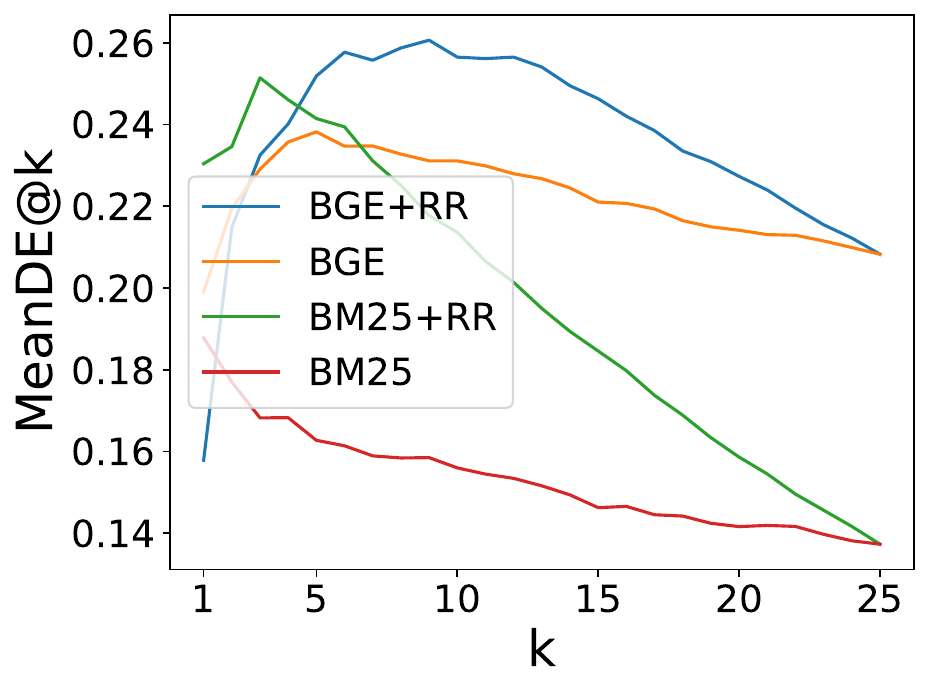}}
    \caption{Results on NQ of different retrieval pipelines when varying the number $k$ of retrieved passages. We compute the distracting effect on Qwen 2.5 7B.}
    \label{fig:retrieval_performance_on_nq}
\end{figure*}

\begin{figure*}
    \centering
    \subfigure[]{\includegraphics[width=0.24\textwidth]{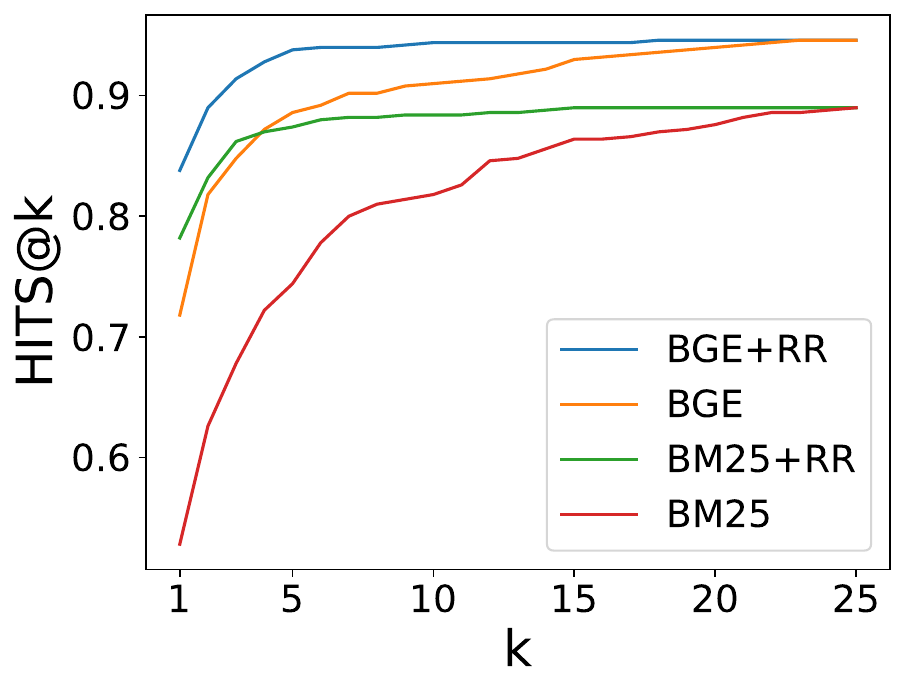}} 
    \subfigure[]{\includegraphics[width=0.24\textwidth]{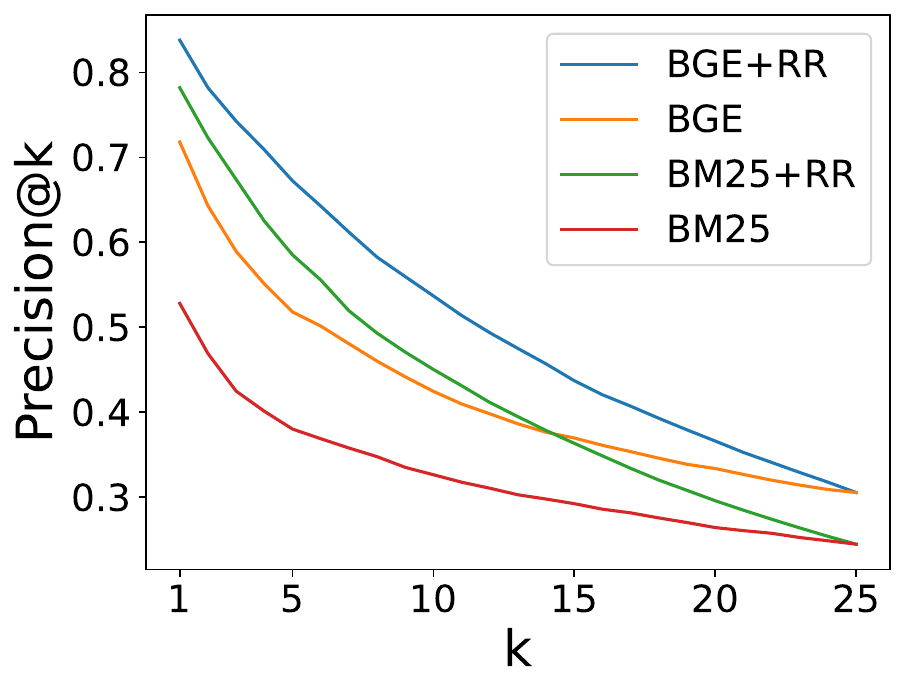}} 
    \subfigure[]{\includegraphics[width=0.24\textwidth]{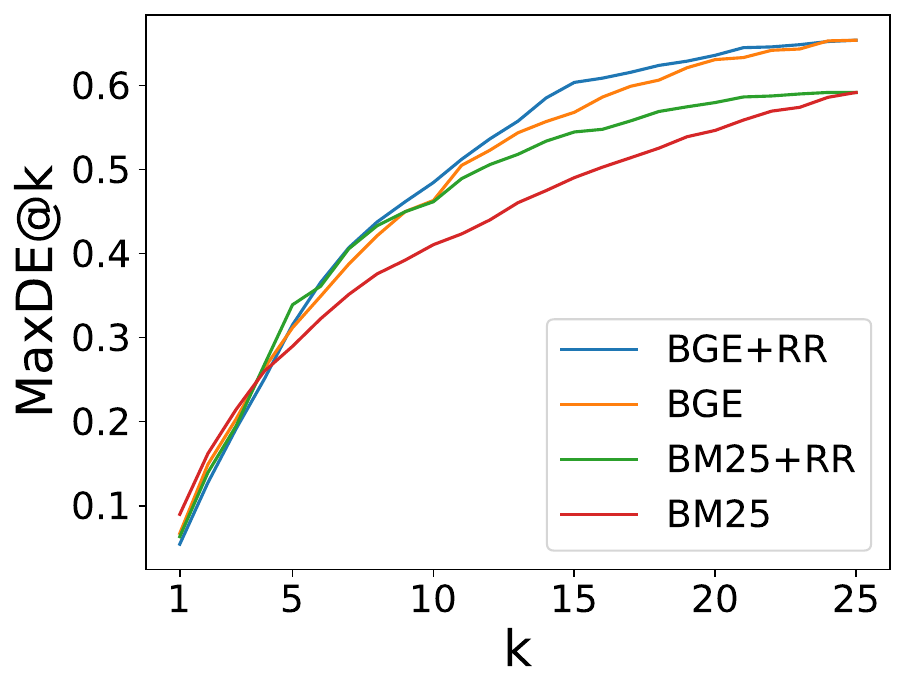}}
    \subfigure[]{\includegraphics[width=0.24\textwidth]{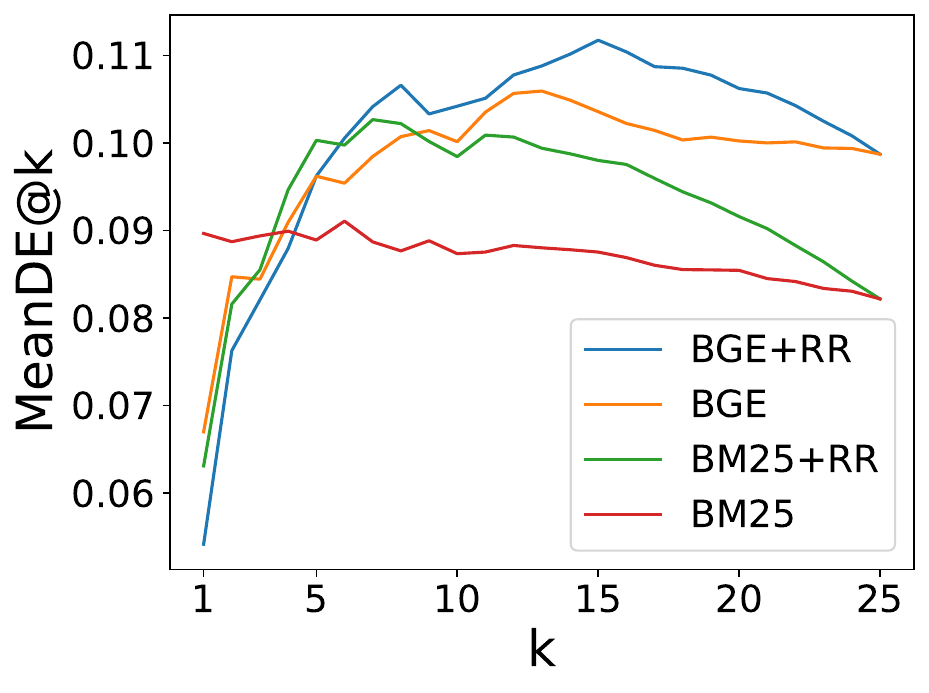}}
    \caption{Results on TriviaQA of different retrieval pipelines when varying the number $k$ of retrieved passages. We compute the distracting effect on Qwen 2.5 7B.}
    \label{fig:retrieval_performance_on_triviaQA}
\end{figure*}

\Cref{fig:retrieval_performance_on_POPQA,fig:retrieval_performance_on_nq,fig:retrieval_performance_on_triviaQA} report the retrieval results of BM25 and BGE with and without re-ranker (RR) on PopQA, NQ, and TriviaQA, respectively. 

Moreover, Table \ref{tab:hard_distractor_prevalence} shows the percentage of queries that contain at least one hard distractor among the top-$k$ retrieved passages. We define a hard distractor as any irrelevant passage with a distracting effect greater than 0.8.

\begin{table}[th]
\centering
\resizebox{\columnwidth}{!}{
\begin{tabular}{llccccc}
\toprule
\multirow{2}{*}{\textbf{Retriever}} & \multirow{2}{*}{\textbf{Benchmark}} & \multicolumn{5}{c}{\textbf{$k$}} \\
\cmidrule{3-7}
& & \textbf{5} & \textbf{10} & \textbf{15} & \textbf{20} & \textbf{25} \\
\midrule
\multirow{4}{*}{BGE + RR} & NQ & 60.60 & 76.00 & 81.20 & 83.00 & 84.20 \\
& TriviaQA & 29.20 & 44.60 & 56.20 & 59.40 & 61.40 \\
& PopQA & 68.40 & 76.00 & 79.60 & 81.20 & 82.60 \\
& \textit{Average} & 52.73 & 65.53 & 72.33 & 74.53 & 76.07 \\
\midrule
\multirow{4}{*}{BGE} & NQ & 58.40 & 73.20 & 77.20 & 82.00 & 84.20 \\
& TriviaQA & 28.00 & 42.60 & 53.20 & 59.20 & 61.40 \\
& PopQA & 63.00 & 72.60 & 76.00 & 80.60 & 82.60 \\
& \textit{Average} & 49.80 & 62.80 & 68.80 & 73.93 & 76.07 \\
\midrule
\multirow{4}{*}{BM25 + RR} & NQ & 56.60 & 68.60 & 71.00 & 72.20 & 72.80 \\
& TriviaQA & 31.40 & 42.20 & 49.80 & 53.40 & 54.40 \\
& PopQA & 59.80 & 68.00 & 71.00 & 71.80 & 72.40 \\
& \textit{Average} & 49.27 & 59.60 & 63.93 & 65.80 & 66.53 \\
\midrule
\multirow{4}{*}{BM25} & NQ & 39.80 & 55.40 & 63.20 & 68.80 & 72.80 \\
& TriviaQA & 25.80 & 36.60 & 44.80 & 50.00 & 54.40 \\
& PopQA & 45.80 & 59.60 & 66.60 & 69.80 & 72.40 \\
& \textit{Average} & 37.13 & 50.53 & 58.20 & 62.87 & 66.53 \\
\bottomrule
\end{tabular}
}
\caption{Percentage of queries having at least one hard distractor in the top-$k$ retrieved passages.}
\label{tab:hard_distractor_prevalence}
\end{table}

\subsubsection{Multi-Vector Retrieval with ColBERT}
We also evaluated ColBERT \citep{khattab2020colbert}, a multi-vector retrieval method that performs fine-grained token-level matching between queries and passages. In terms of retrieval quality metrics (HITS@$k$, Precision@$k$, MaxDE@$k$, MeanDE@$k$), ColBERT exhibits trends similar to BGE (see Fig. \ref{fig:retrieval_performance}). For the passage arrangement experiments in Section \ref{sec:real}, ColBERT achieved the following accuracies with Qwen 2.5 7B on the top-5 retrieved passages: \texttt{Sequential} (67.13), \texttt{Inverse} (67.20), \texttt{Shuffle} (67.00), \texttt{MaxRelevance} (66.87), and \texttt{MinDistraction} (67.07). These results show no statistically significant differences between positional strategies (Wilcoxon test, $p$=0.05), reinforcing our main conclusion that even sophisticated multi-vector retrievers surface a mixture of relevant and highly distracting passages, thereby mitigating positional bias effects in practical RAG scenarios.

\subsection{LLM-as-a-Judge Methodology} \label{sec:llm_as_judge}
A critical aspect of our work is the reliable classification of passages as relevant or irrelevant.  We placed particular emphasis on minimizing false negatives, i.e., passages incorrectly labeled as irrelevant despite containing useful information to answer the question. Therefore, we employed a strong LLM, namely Claude 3.7 Sonnet, to judge if a passage is relevant or not. We prompted the LLM to evaluate relevance by considering the question, the passage, the ground truth answers from the dataset, and few-shot examples as demonstrations of relevant and irrelevant passages, with a particular focus on distracting passages. The exact prompt is shown in Fig. \ref{fig:relevance_prompt}.

For answer quality evaluation, we prompted the same LLM to assess whether the generated response semantically matches reference answers. This approach prevents penalizing correct answers that use different phrasing than the reference, ensuring our effectiveness metrics genuinely reflect the model's ability to extract and utilize information rather than simply mimic exact answer formats. For example, if the ground truth answer to ``What is the population of Yokyo?'' is ``14 million people'', a generated answer like ``14 million residents'' would be correctly judged as semantically equivalent under our evaluation approach, while it would be considered incorrect under classical exact match metrics.
We took inspiration from the OpenAI template used in \citet{wei2024simpleqa}, with modifications to adapt to our specific task requirements. 
Fig. \ref{fig:answer_correctness_prompt} provides the exact prompt used for answer quality assessment.

\subsubsection{Open-Source Alternative Validation}
\label{sec:llm_judge_comparison}
We also tested whether strong open-source models can serve as reliable alternatives to Claude 3.7 Sonnet for our evaluation tasks. We conducted a comparative analysis using Llama 3.3 70B Instruct on a subset of our data.
We sampled 100 queries from each dataset (300 total) and evaluated relevance decisions for the top-5 passages retrieved by BGE using both models, analyzing a total of 1500 passages. For passage relevance assessment, the Cohen's Kappa coefficient between Claude and Llama yielded a score of 0.85, indicating substantial agreement according to standard interpretation guidelines \cite{landis1977measurement}.
For answer accuracy evaluation, we had both models assess the correctness of 300 generated answers. The agreement was even stronger, with a Cohen's Kappa score of 0.94, indicating almost perfect agreement.
These results suggest that Llama 3.3 70B Instruct can serve as a reliable open-source alternative for both passage relevance assessment and answer quality evaluation in RAG experiments.

\section{Results for Other LLMs and Single Datasets}
\label{sec:results_other_llms}
In this section, we present detailed results for all LLMs and individual datasets. While the main paper reported results averaged across datasets for space constraints, here we analyze the positional bias effects for each dataset and different LLMs.

% \begin{table}[t]
% \centering
% \small
% % \resizebox{\columnwidth}{!}{
% \begin{tabular}{lccc}
% \textbf{LLM} & \textbf{NQ} & \textbf{TriviaQA} & \textbf{PopQA} \\ \hline
% Q7B          & 44.20       & 68.80             & 20.40          \\
% L3B          & 58.60       & 68.00             & 20.60          \\
% L8B          & 67.40       & 80.80             & 30.60          \\
% L70B         & 74.60       & 92.20             & 49.60         
% \end{tabular}
% % }
% \vspace{-0.2cm}
% \caption{Closed-book answer accuracy for different LLMs across the three benchmarks.}
% \label{tab:closed_book}
% \end{table}

\begin{table}[b]
\centering
\small
\begin{tabular}{@{}lccc@{}}
\toprule
\textbf{LLM} & \textbf{NQ} & \textbf{TriviaQA} & \textbf{PopQA} \\ \midrule
Q7B          & 44.20       & 68.80             & 20.40          \\
L3B          & 58.60       & 68.00             & 20.60          \\
L8B          & 67.40       & 80.80             & 30.60          \\
L70B         & 74.60       & 92.20             & 49.60          \\ \bottomrule
\end{tabular}
\caption{Closed-book answer accuracy for different LLMs across the three benchmarks.}
\label{tab:closed_book}
\end{table}

\subsection{Positional Bias in Controlled Settings}
\begin{figure*}[t!]
    \centering
    \includegraphics[width=\linewidth]{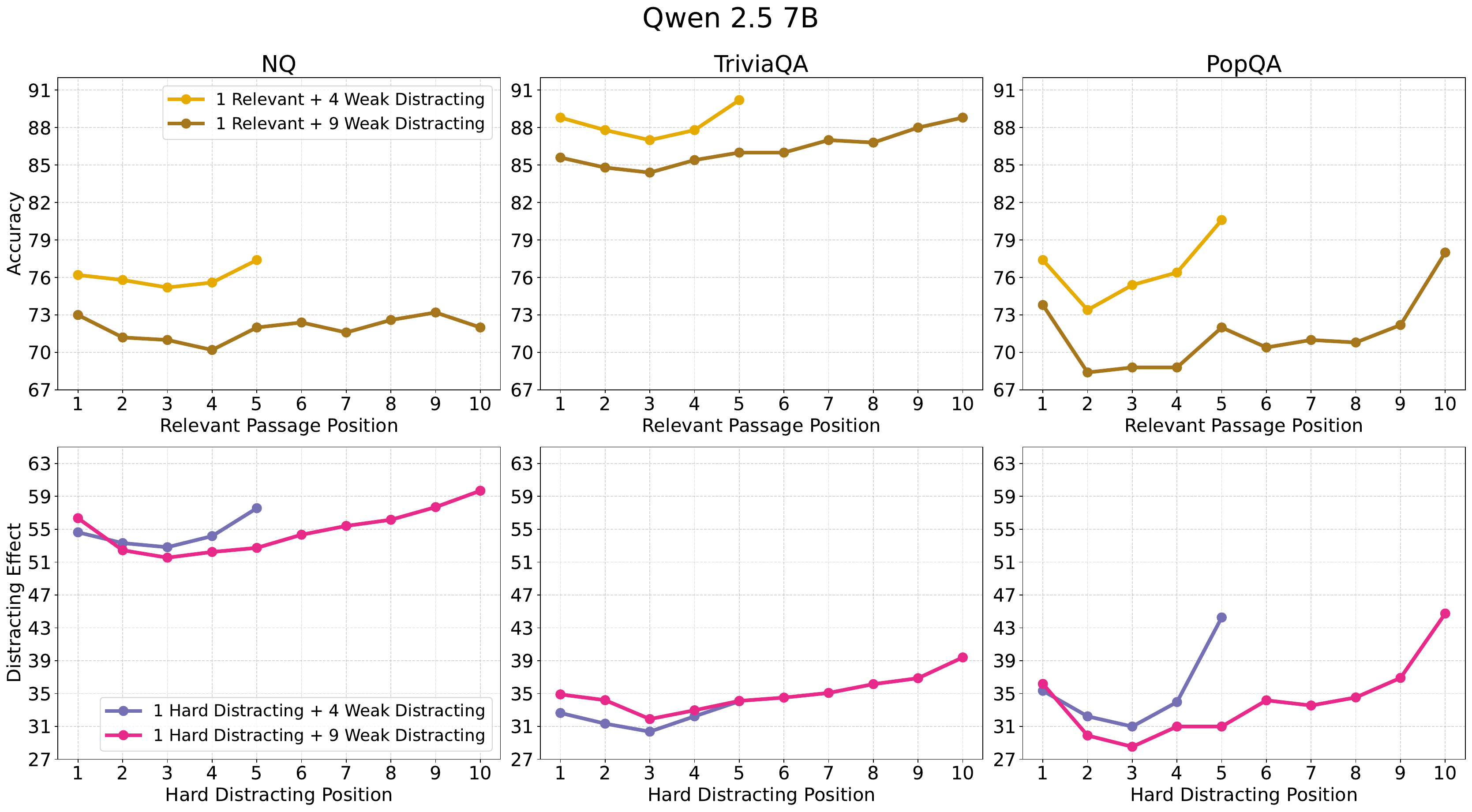}
    \caption{Controlled experiments results for Qwen 2.5 7B across datasets.}
    \label{fig:qwen7b_position_analysis}
\end{figure*}

\begin{figure*}[t!]
    \centering
    \includegraphics[width=\linewidth]{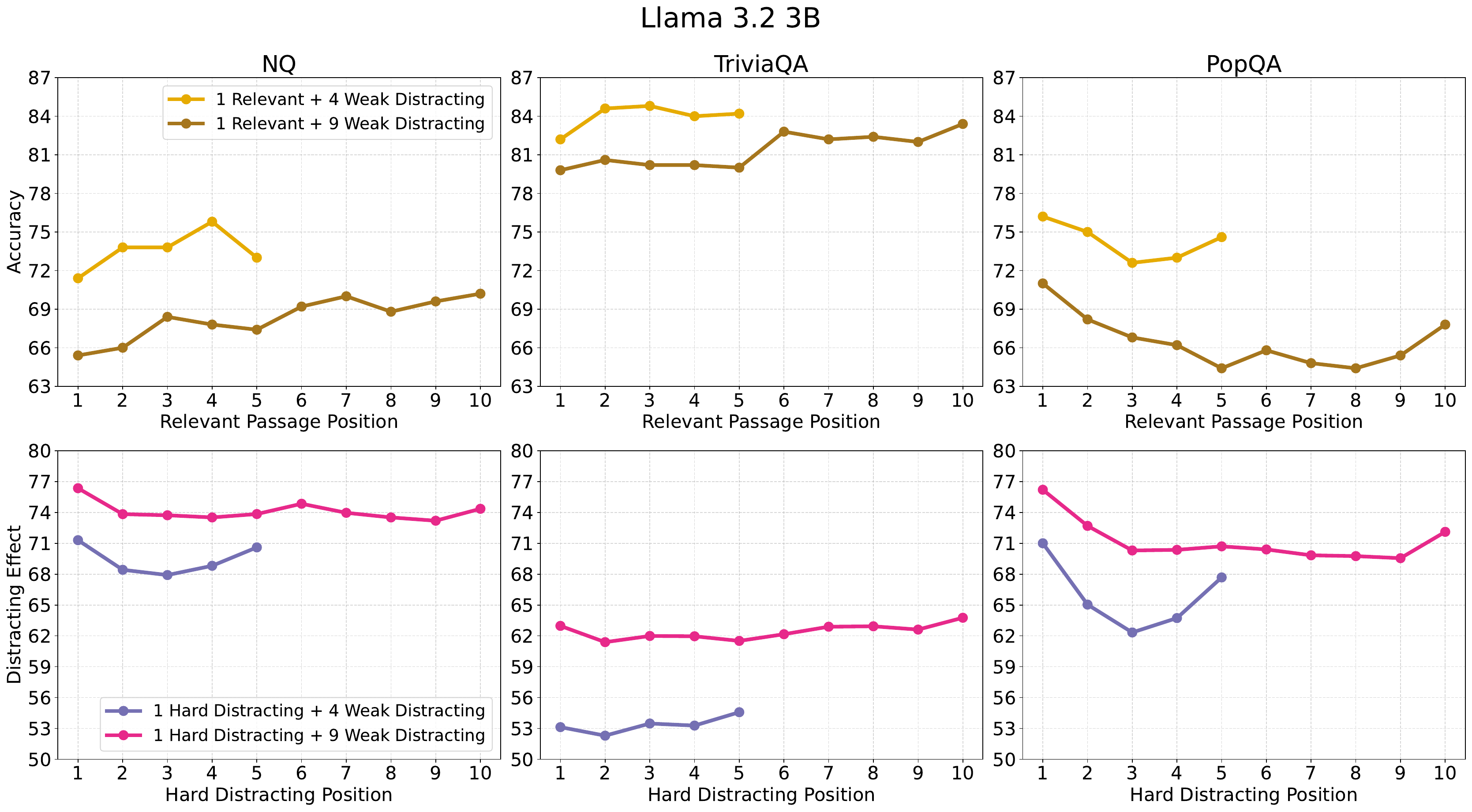}
    \caption{Controlled experiments results for Llama 3.2 3B across datasets.}
    \label{fig:llama3b_position_analysis}
\end{figure*}

\begin{figure*}[t!]
    \centering
    \includegraphics[width=\linewidth]{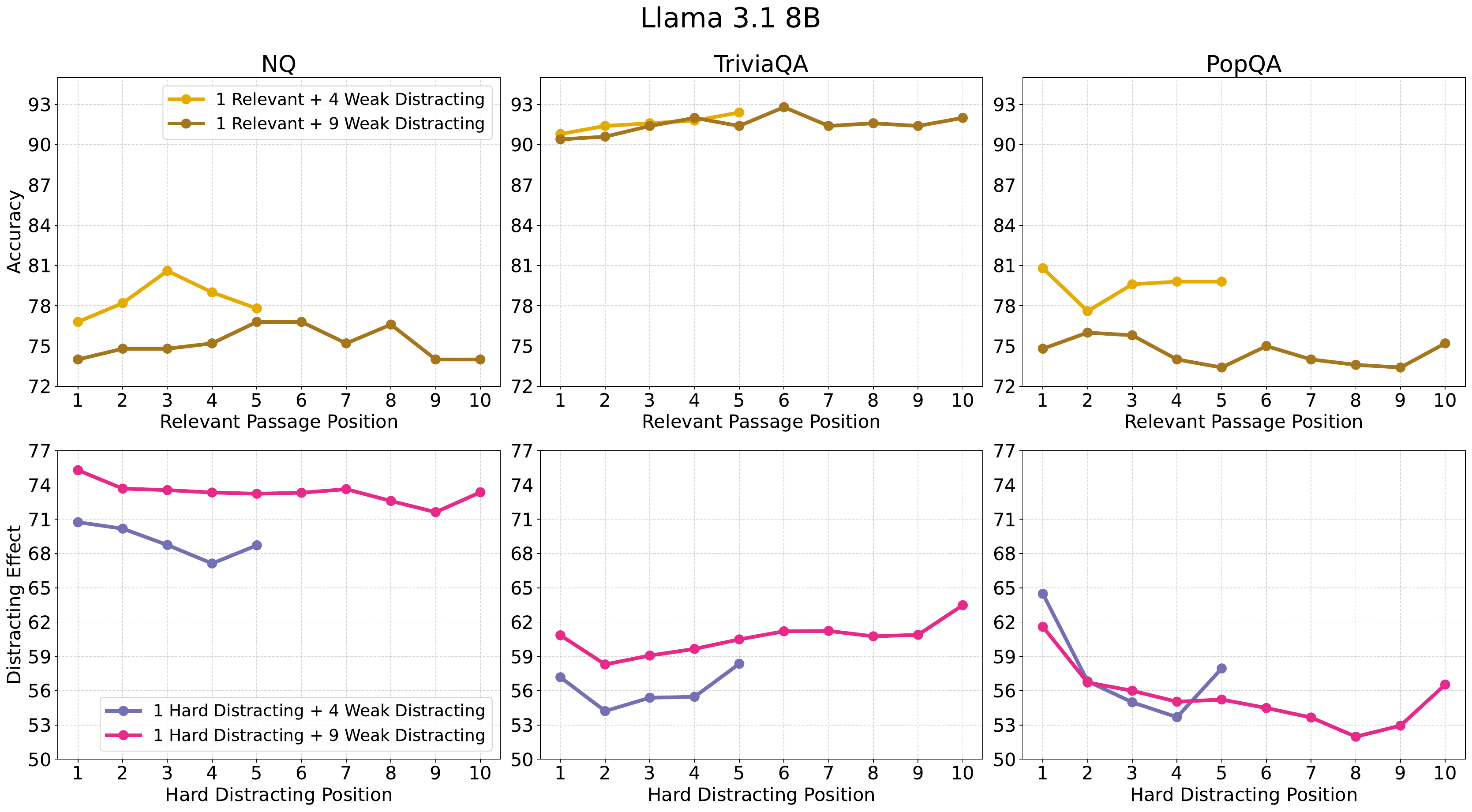}
    \caption{Controlled experiments results for Llama 3.1 8B across datasets.}
    \label{fig:llama8b_position_analysis}
\end{figure*}

\begin{figure*}[t!]
    \centering
    \includegraphics[width=\linewidth]{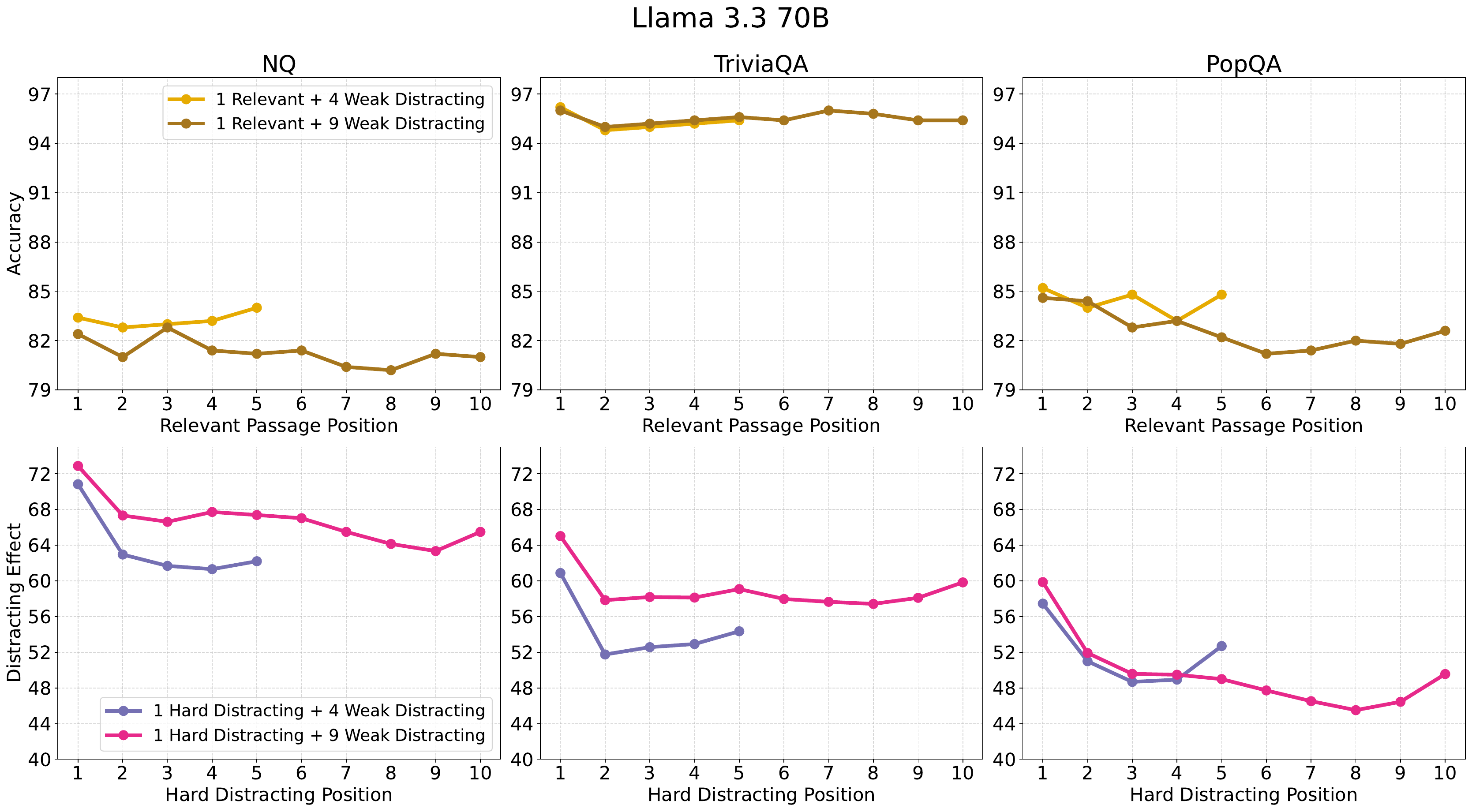}
    \caption{Controlled experiments results for Llama 3.3 70B across datasets.}
    \label{fig:llama70b_position_analysis}
\end{figure*}

\Cref{fig:qwen7b_position_analysis,fig:llama3b_position_analysis,fig:llama8b_position_analysis,fig:llama70b_position_analysis} illustrate the positional bias in controlled settings when rotating either the relevant passage or a hard distractor among weak distractors. The results reveal that each model exhibits its own characteristic positional pattern, confirming findings from \citet{hutter-2025-lost-but-not-only}.

Among the LLMs tested, Qwen 2.5 7B demonstrates the most pronounced positional bias (see Fig. \ref{fig:qwen7b_position_analysis}), while the Llama 3 family appears more resilient to position changes (see \Cref{fig:llama3b_position_analysis,fig:llama8b_position_analysis,fig:llama70b_position_analysis}). A possible explanation is that these models may have been specifically trained to mitigate the \textit{lost-in-the-middle} effect. Since this problem has become well-documented in the literature \citep{liu-etal-2024-lost, he-etal-2024-never, hsieh2024found}, Llama models might incorporate architectural modifications or training techniques designed to maintain robust attention across all positions in the context window, making them less susceptible to passage positioning issues.

In addition, this different behavior from positional bias can be further explained by examining the closed-book effectiveness of these models (Table~\ref{tab:closed_book}).  For the KILT benchmarks, models like Llama 3.3 70B achieve remarkably high closed-book accuracy (74.60 for NQ and 92.20 on TriviaQA), suggesting extensive memorization during pretraining. When LLMs encounter questions they already know the answer to, they tend to rely on their parametric knowledge rather than context, especially when the relevant passage appears in a non-preferential position. This \emph{parametric bias} has been observed by \citet{kortukov2024studying}, who found that LLMs' factual parametric knowledge can negatively influence their reading abilities and behaviors, leading to a preference for known information over contextual evidence.

This pattern differs for PopQA, where closed-book accuracy is significantly lower across all models. PopQA contains questions about long-tail entities that are less represented in the models' parametric memory \citep{mallen2023not}, making contextual information more crucial. For smaller models (Llama 3.2 3B, Llama 3.1 8B, and Qwen 2.5 7B), PopQA exhibits stronger positional effects when rotating the relevant passage. The effect is less pronounced in Llama 3.3 70B due to its larger parametric memory that can often recall these long-tail entities.

Regarding distracting effects, when rotating a hard distractor among weak distracting passages, all models generally display the characteristic U-shaped pattern (see \Cref{fig:qwen7b_position_analysis,fig:llama3b_position_analysis,fig:llama8b_position_analysis,fig:llama70b_position_analysis}), suggesting that distracting effects are more consistent across models and less influenced by parametric knowledge.

% \begin{table}[t]
% \setlength{\tabcolsep}{3.2pt}
% \footnotesize
% \begin{tabular}{lccccc}
% \textbf{LLM} & \textbf{Sequential} & \textbf{Inverse} & \textbf{Shuffle} & \textbf{MaxRel} & \textbf{MinDist} \\ \hline
% Q7B  & 70.20 & 71.00 & 71.40 & 71.33 & 70.33 \\
% L3B  & 64.47 & 66.47 & 65.67 & 65.80 & 65.73 \\
% L8B  & 68.47 & 70.80 & 70.07 & 68.80 & 69.00 \\
% L70B & 75.13 & 75.00 & 75.67 & 76.13 & 74.33
% \end{tabular}
% \vspace{-0.2cm}
% \caption{Answer accuracy for different LLMs when arranging the top-10 passages retrieved by BGE+RR using different strategies.}
% \label{tab:all_llms_res_k_10}
% \end{table}

\begin{table}[t]
\setlength{\tabcolsep}{3.2pt}
\footnotesize
\begin{tabular}{@{}lccccc@{}}
\toprule
\textbf{LLM} & \textbf{Sequential} & \textbf{Inverse} & \textbf{Shuffle} & \textbf{MaxRel} & \textbf{MinDist} \\ \midrule
Q7B  & 70.20 & 71.00 & 71.40 & 71.33 & 70.33 \\
L3B  & 64.47 & 66.47 & 65.67 & 65.80 & 65.73 \\
L8B  & 68.47 & 70.80 & 70.07 & 68.80 & 69.00 \\
L70B & 75.13 & 75.00 & 75.67 & 76.13 & 74.33 \\ \bottomrule
\end{tabular}
\caption{Answer accuracy for different LLMs when arranging the top-10 passages retrieved by BGE+RR using different strategies.}
\label{tab:all_llms_res_k_10}
\end{table}

\subsection{Positional Bias in Real Scenarios}\label{sec:app_real_setting}

In Section~\ref{sec:real}, we presented experiments for $k$=5, showing minimal impact of different passage arrangement strategies on answer accuracy. Here, we expand the analysis to $k$=10 to investigate whether retrieving more passages might exhibit a more pronounced positional bias effect.
Table~\ref{tab:all_llms_res_k_10} shows the answer accuracy across different LLMs when arranging the top-10 passages retrieved by BGE+RR using the strategies described in Section~\ref{sec:real}. Similar to the $k$=5 case, we observe that the positional bias has a marginal impact on answer accuracy. Across all LLMs, the difference between the best-performing strategy and the \texttt{Shuffle} strategy is not statistically significant according to the Wilcoxon test with $p$-value=0.05. 
Table~\ref{tab:all_pipelines_res_k_10} presents the results for Qwen 2.5 7B across different retrieval pipelines. We note one exception in the BGE retriever (without re-ranker), where the \texttt{MaxRelevance} strategy achieves 71.00 accuracy while \texttt{Shuffle} yields 68.87, which is a statistically significant difference. However, this appears to be an isolated case rather than a consistent pattern. This single exception does not contradict the broader statistical trend observed across all other configurations.
For weaker retrievers like BM25, the positional ordering has less impact simply because they retrieve fewer relevant passages overall, as shown in Fig.~\ref{fig:retrieval_performance}a.

In general, these findings with $k$=10 reinforce our conclusion from the main paper: in realistic RAG settings, the impact of positional bias is minimal compared to its effect in controlled experimental conditions. The interaction between relevant and distracting passages in real retrieval results tends to neutralize potential benefits from strategic passage ordering.

% \begin{table}[t]
% \setlength{\tabcolsep}{1.6pt}
% \footnotesize
% \begin{tabular}{lccccc}
% \textbf{Retriever} & \textbf{Sequential} & \textbf{Inverse} & \textbf{Shuffle} & \textbf{MaxRel} & \textbf{MinDist} \\ \hline
% BGE     & 69.33 & 69.73 & 68.87 & 71.00 & 68.40 \\
% BGE+RR  & 70.20 & 71.00 & 71.40 & 71.33 & 70.33 \\
% BM25    & 54.73 & 54.60 & 55.93 & 56.07 & 55.00 \\
% BM25+RR & 59.93 & 60.60 & 60.40 & 60.73 & 59.07
% \end{tabular}
% \vspace{-0.2cm}
% \caption{Answer accuracy of Qwen 2.5 7B when arranging with different strategies the top-10 passages retrieved from different retrieval pipelines.}
% \label{tab:all_pipelines_res_k_10}
% \end{table}

\begin{table}[t]
\setlength{\tabcolsep}{1.6pt}
\footnotesize
\begin{tabular}{@{}lccccc@{}}
\toprule
\textbf{Retriever} & \textbf{Sequential} & \textbf{Inverse} & \textbf{Shuffle} & \textbf{MaxRel} & \textbf{MinDist} \\ \midrule
BGE     & 69.33 & 69.73 & 68.87 & 71.00 & 68.40 \\
BGE+RR  & 70.20 & 71.00 & 71.40 & 71.33 & 70.33 \\
BM25    & 54.73 & 54.60 & 55.93 & 56.07 & 55.00 \\
BM25+RR & 59.93 & 60.60 & 60.40 & 60.73 & 59.07 \\ \bottomrule
\end{tabular}
\caption{Answer accuracy of Qwen 2.5 7B when arranging with different strategies the top-10 passages retrieved from different retrieval pipelines.}
\label{tab:all_pipelines_res_k_10}
\end{table}

\clearpage
\newpage

\begin{figure*}
\begin{mdframed}[font=\footnotesize]
\begin{Verbatim}[breaklines=true, breaksymbol=]
You are given a question and you must respond based on the provided documents. Respond directly without providing any premise or explanation.

Documents:
<passage>
...
<passage>

Question: 
<question>

Answer:
\end{Verbatim}
\end{mdframed}
\vspace{-0.4cm}
\caption{Prompt used for response generation.} \label{fig:prompt dist test}
\end{figure*}

\begin{figure*}
\begin{mdframed}[font=\tiny, 
                 innertopmargin=2pt,    % Default is 4pt
                 innerbottommargin=2pt, % Default is 4pt
                 skipabove=0pt,         % Space above frame
                 skipbelow=0pt] 
\begin{Verbatim}[breaklines=true, breaksymbol=]
Your job is to look at a question, a list of acceptable answers, and a document, then determine if the document is RELEVANT or IRRELEVANT for answering the question. Each document may have some metadata information like the title or the section it belongs to. This information may help you understand the context of the document. We are in a multi-reference setting, which means that there may be multiple correct answers to the question. The answer list contains all the correct answers.

First, I will give examples of each type, and then you will evaluate a new example.
The following are examples of RELEVANT documents.
```
Question 1: when did korn's follow the leader come out
Acceptable answers list 1: ['August 18 , 1998', 'Summer 1998']
Document 1: (Title: Follow the Leader (Korn album)) Follow the Leader is the third studio album by the American nu metal band Korn . The album was released on August 18 , 1998 , through Immortal / Epic . This was their first album not produced by Ross Robinson . Instead , it was produced by Steve Thompson and Toby Wright .

Question 2: who played bobby byrd in get on up
Acceptable answers list 2: ['Nelsan Ellis']
Document 2: (Title: Get on Up (film))(Section: Production - Casting) On August 26, 2013, Universal selected Chadwick Boseman to play the lead role of James Brown. Boseman did all of his own dancing and some singing. The soundtrack is live recordings of James Brown. On September 17, Universal announced an open casting call for actors, musicians, and extras for different roles in the biopic, which was held on September 21. On September 30, Taylor cast Viola Davis to play Susie Brown and Octavia Spencer to play Aunt Honey. On October 21, Nelsan Ellis joined the cast of film to portray Bobby Byrd, Brown's long-time friend.

Question 3: What movie has the song on the road again?
Acceptable answers list 3: ['Honeysuckle Rose']
Document 3: (Title: On the Road Again (Willie Nelson song)) The song , about life on tour , came about when the executive producer of the film Honeysuckle Rose approached Nelson about writing the song for the film 's soundtrack . '' On the Road Again '' became Nelson 's 9th Country & Western No. 1 hit overall ( 6th as a solo recording act ) in November 1980 , and became one of Nelson 's most recognizable tunes . In addition , the song reached No. 20 on the Billboard Hot 100 , and No. 7 on the Adult Contemporary chart . It was his biggest pop hit to that time and won him a Grammy Award for Best Country Song a year later .
```
These documents are all RELEVANT because:
    - They contain sufficient information to support at least ONE of the acceptable answers.
    - The information can be found directly or through simple inference.
    - Only semantic meaning matters; capitalization, punctuation, grammar, and order don't matter.

The following are examples of IRRELEVANT documents.
```
Question 1: when did korn's follow the leader come out
Acceptable answers list 1: ['August 18 , 1998', 'Summer 1998']
Document 1: (Title: Korn Discography) Korn's third album marked a significant evolution in their sound and commercial success. The band spent much of 1998 recording and promoting this album, which would go on to achieve platinum status multiple times. Following their summer tour, they continued to gain mainstream attention. The album contained several singles that performed well on the charts, including "Got the Life" and "Freak on a Leash." Reviews were generally positive, with critics noting the band had refined their nu-metal style while maintaining their aggressive edge.

Question 2: who played bobby byrd in get on up
Acceptable answers list 2: ['Nelsan Ellis']
Document 2: (Title: Get on Up (film))(Section: Critical Reception) Critics particularly praised the casting decisions in "Get on Up," noting the strong ensemble supporting Chadwick Boseman's portrayal of James Brown. The film's recreation of the dynamic between Brown and his longtime friend and musical collaborator received significant attention. Several reviewers highlighted the chemistry between the main characters and how it captured their complex professional and personal relationship spanning decades. The scenes depicting their early musical development were considered among the film's strongest moments, effectively showing how their partnership shaped the evolution of funk music.

Question 3: What movie has the song on the road again?
Acceptable answers list 3: ['Honeysuckle Rose']
Document 3: (Title: Classic Songs in Films) Many people believe, though it's not actually correct, that Willie Nelson's iconic song 'On The Road Again' first appeared in the 1980 film 'Smokey and the Bandit II.' Some music historians have suggested that this misconception arose because the film's themes of truck driving and life on the road seemed to perfectly match the song's message. The song's road-trip vibe made it a natural fit for many movies, but this particular connection is just a popular misconception.
```
These documents are all IRRELEVANT because:
    - They lack the necessary information to support any of the acceptable answers, even though they may contain some related information.
    - They reference similar themes, keywords, or surrounding context but don't provide the specific answer required.
    - Some contain subtle distractors that seem relevant at first glance but don't actually answer the specific question.

Before making your final evaluation, follow this step-by-step process:
1. Identify the specific information needed to match at least one of the acceptable answers.
2. Carefully search the document for this exact information or information that directly implies it.
3. Check for these common errors:
   - The document contains similar keywords or themes but not the actual answer.
   - The document contains partial information that would need to be combined with external knowledge.
   - The document discusses related topics but doesn't specifically answer the question.
Also note the following things:
- The evaluation should be based ONLY on the specific question and acceptable answers list provided.
- Do not try to generalize or apply your own knowledge beyond the information given in the question, acceptable answers list, and document.
- A document with tangential information about the topic is still IRRELEVANT if it doesn't contain the specific answer.

Here is a new example. Don't apologize or correct yourself if there was a mistake; we are just trying to evaluate the relevance of the document.
```
Question: {question}
Acceptable answers list: {answers}
Document: {document}
```
Evaluate the document for this new question as one of:
A: RELEVANT
B: IRRELEVANT

Return a JSON object with the following format:
{{
    "motivation": "Your concise motivation for the evaluation here. Use maximum 2 sentences.",
    "grade": "A" or "B"
}}
\end{Verbatim}
\end{mdframed}
% \vspace{-0.4cm}
\caption{Prompt for document relevance assessment using Claude 3.7 Sonnet as judge.} 
\label{fig:relevance_prompt}
\end{figure*}

\begin{figure*}
\begin{mdframed}[font=\tiny]
\begin{Verbatim}[breaklines=true, breaksymbol=]
Your job is to look at a question, a list of acceptable answers, and a predicted answer, and then assign a grade of either CORRECT or INCORRECT.
We are in a multi-reference setting, which means that there may be multiple correct answers to the question. The answer list contains all the correct answers.

First, I will give examples of each grade, and then you will grade a new example.

The following are examples of CORRECT predicted answers.
```
Question: What are the names of Barack Obama's children?
Acceptable answers list: ['Malia Obama and Sasha Obama', 'Natasha Marian and Malia Ann']
Predicted answer 1: sasha and malia obama
Preidcted answer 2: Natasha and Malia
Predicted answer 3: most people would say Malia and Sasha, but I'm not sure and would have to double check
Predicted answer 4: Barack Obama has two daughters. Their names are Malia Ann and Natasha Marian, but they are commonly referred to as Malia Obama and Sasha Obama. Malia was born on July 4, 1998, and Sasha was born on June 10, 2001.
```
These predicted answers are all CORRECT because:
    - They contain all essential information from at least one of the acceptable answers.
    - They do not contain any information that contradicts the acceptable answers.
    - Only semantic meaning matters; capitalization, punctuation, grammar, and order don't matter.
    - Hedging and guessing are permissible, provided that at least one of the acceptable answers is fully included and the response contains no incorrect information or contradictions.


The following are examples of INCORRECT predicted answers.
```
Question: What are the names of Barack Obama's children?
Acceptable answers list: ['Malia and Sasha', 'Natasha Marian and Malia Ann']
Predicted answer 1: Malia.
Predicted answer 2: Malia, Sasha, and Susan.
Predicted answer 3: Barack Obama does not have any children.
Predicted answer 4: I think it's either Malia and Sasha. Or it could be Malia and Jackie. Or it could be Joey and Malia.
Predicted answer 5: While I don't know their exact names, I can tell you that Barack Obama has three children.
Predicted answer 6: It's possible you may mean Betsy and Olivia. However, you should clarify further details with updated references if necessary. Is that the correct answer?
Predicted answer 7: It may be the case that Obama's child is named James. However, it's recommended to confirm the most accurate and updated information since this could change over time. This model may not always reflect the most current information.
Predicted answer 8: Malia and Sasha are Barack Obama's daughters. Additionally, he adopted a son named Michael in 2018.
```
These predicted answers are all INCORRECT because:
    - A factual statement in the answer contradicts the information in the acceptable answers list. Incorrect statements that have some hedging (e.g., "it is possible that", "although i'm not sure, i think") are also considered incorrect.
    In particular:
        - Predicted answer 1: It only mentions one child when both are required.
        - Predicted answers 2, 5, and 8: They include accurate information but also contain incorrect information ("Susan", "three children", "adopted a son named Michael").
        - Predicted answer 3: It directly contradicts the acceptable answer.
        - Predicted answer 4: It presents multiple alternatives, some of which are incorrect ("Jackie", "Joey").
        - Predicted answers 6 and 7: They suggest completely incorrect answers, even with hedging ("Betsy and Olivia", "James").


Also note the following things:
- The acceptable answers may contain more information than the question. In such cases, the predicted answer only needs to contain the information that is in the question.
    - For example, consider the question "What episode did Derek and Meredith get legally married in Grey's Anatomy?" with acceptable answer "Season 7, Episode 20: White Wedding". Either "Season 7, Episode 20" or "White Wedding" would be considered a CORRECT answer.
- Do not punish predicted answers if they omit information that would be clearly inferred from the question.
    - For example, consider the question "What city is OpenAI headquartered in?" and the acceptable answer "San Francisco, California". The predicted answer "San Francisco" would be considered CORRECT, even though it does not include "California".
    - Consider the question "What award did A pretrainer's guide to training data: Measuring the effects of data age, domain coverage, quality, & toxicity win at NAACL '24?", the acceptable answer is "Outstanding Paper Award". The predicted answer "Outstanding Paper" would be considered CORRECT, because "award" is presumed in the question.
    - For the question "What is the height of Jason Wei in meters?", the acceptable answer is "1.73 m". The predicted answer "1.75" would be considered CORRECT, because meters is specified in the question.
    - For the question "What is the name of Barack Obama's wife?", the acceptable answer is "Michelle Obama". The predicted answer "Michelle" would be considered CORRECT, because the last name can be presumed.
- Do not punish for typos in people's name if it's clearly the same name.
    - For example, if the acceptable answer is "Hyung Won Chung", you can consider the following predicted answers as correct: "Hyoong Won Choong", "Hyungwon Chung", or "Hyun Won Chung".
- You must attain to the question and acceptable answers list information. Do not rely on your knowledge to grade the answer. If the predicted answer is correct based on the question and acceptable answer, it should be considered correct, even if you know that the answer is wrong.


Here is a new example. Don't apologize or correct yourself if there was a mistake; we are just trying to grade the answer.
```
Question: {question}
Acceptable answers list: {target}
Predicted answer: {predicted_answer}
```

Grade the predicted answer of this new question as one of:
A: CORRECT
B: INCORRECT

Return a JSON object with the following format:
{{
    "motivation": "Your concise motivation for the grade here. Use maximum 2 sentences.",
    "grade": "A" or "B"
}}
\end{Verbatim}
\end{mdframed}
\caption{Prompt for answer correctness assessment using Claude 3.7 Sonnet as judge.} 
\label{fig:answer_correctness_prompt}
\end{figure*}

\end{document}